\documentclass{article}

\usepackage{arxiv}

\usepackage[utf8]{inputenc} 
\usepackage{hyperref}       
\usepackage{url}            
\usepackage{booktabs}       
\usepackage{amsfonts}       
\usepackage{nicefrac}       
\usepackage{microtype}      
\usepackage{graphicx}
\usepackage{natbib}
\usepackage{doi}

\usepackage{bibunits}

\defaultbibliography{references}
\defaultbibliographystyle{unsrtnat}

\usepackage{amsmath}
\usepackage{algpseudocode,algorithm,algorithmicx}

\usepackage{float}

\usepackage{lmodern}
\usepackage{siunitx}
\usepackage{booktabs}
\usepackage{etoolbox}

\usepackage[margin=20pt]{subfig}
\usepackage{amssymb}
\usepackage{amsmath}
\usepackage[cmyk]{xcolor}  
\usepackage{colortbl}
\usepackage{arydshln}
\usepackage{multirow}

\definecolor{c1}{HTML}{003f5c}
\definecolor{c2}{HTML}{444e86}
\definecolor{c3}{HTML}{955196}
\definecolor{c4}{HTML}{dd5182}
\definecolor{c5}{HTML}{ff6e54}
\definecolor{c6}{HTML}{ffa600}

\usepackage{soul}

\title{Language Modelling via Learning to Rank}

\date{} 					

\author{%
  Arvid Frydenlund\thanks{Contact author. Work accepted to AAAI 2022.} \\
  University of Toronto\\
  Vector Institute for Artificial Intelligence\\
  \texttt{arvie@cs.toronto.edu} \\
   \And
   Gagandeep Singh \\
   Nuance Communications Inc. \\
   \texttt{gagandeep.singh1@nuance.com} \\
   \And
   Frank Rudzicz \\
  University of Toronto \\
  Vector Institute for Artificial Intelligence\\
   Unity Health Toronto\\
   \texttt{frank@cs.toronto.edu} \\
}

\renewcommand{\shorttitle}{Language Modelling via Learning to Rank}

\hypersetup{
pdftitle={Language Modelling via Learning to Rank},
pdfsubject={NLP, ML},
pdfauthor={Arvid Frydenlund},
pdfkeywords={Language Models, Ranking, Plackett-Luce, Label Smoothing, Knowledge Distillation},
}

\begin{document}

\begin{bibunit}

\maketitle

\begin{abstract}


We consider language modelling (LM) as a multi-label structured prediction task by re-framing training from solely predicting a single ground-truth word to ranking a set of words which could continue a given context. To avoid annotating top-$k$ ranks, we generate them using pre-trained LMs: GPT-2, BERT, and Born-Again models.  This leads to a rank-based form of knowledge distillation (KD). We also develop a method  using  $N$-grams to create a non-probabilistic teacher which generates the ranks without the need of a pre-trained LM.

We confirm the hypotheses that we can treat LMing as a ranking task and that we can do so without the use of a pre-trained LM.
We show that rank-based KD generally improves perplexity (PPL) ---  often with statistical significance --- when compared to Kullback–Leibler-based KD.  Surprisingly, given the simplicity of the method, the $N$-grams act as  competitive teachers and achieve similar performance as using either BERT or a Born-Again model as teachers. GPT-2 always acts as the best teacher, though, and using it and a Transformer-XL student on Wiki-02, rank-based KD reduces a cross-entropy baseline from 65.27 to 55.94 and against a KL-based KD of 56.70.

\end{abstract}

\section{Introduction and motivation}\label{sec::intro}

\noindent  More often than not, there are many ways to say the same thing.  For example, `{\em the cat sat on the mat}' is semantically equivalent to `{\em the cat sat on the rug}', in most contexts.  This basic fact about language is ignored when training language models, where we try to maximize the probability of predicting a single ground-truth word for a given context and penalize all other words regardless of any potential semantic equivalence. In particular, a word-level language model (LM) defines a distribution over $T$ discrete tokens, $x_1, \dots, x_{T}$, as

\begin{equation}\label{eq::LM}
\begin{split}
p(x_1, \dots, x_T \,|\, x_0) & = \prod_{t=1}^T p(x_t \,|\, x_{< t} )   = \prod_{t=1}^T \frac{e^{w_t^{g_t}}}{\sum_{j=1}^{|V|} e^{w_t^j}} = \prod_{t=1}^T \frac{e^{w_t^{g_t}}}{Z_t},
\end{split}
\end{equation}
where we use the softmax function to model a multinomial distribution over a set vocabulary $V$, $w_t \in \mathbb{R}^{|V|}$ are the logits produced by the model at step $t$,  $j$ indexes the score for the $j^{\mathrm{th}}$ word-type, and $g_t$ is the ground-truth (GT) index at time $t$. Such models are trained by minimizing the per-word cross-entropy (CE or $H(y, p)$) 
against the ground-truth text, $y = y_1, \dots, y_T \in \{0, 1\}^{|V|}$ which are one-hot representations of $x_1, \dots, x_T$,

\begin{equation}\label{eq::CE}
\mathrm{CE} = -\frac{1}{T} \sum_{t=1}^T \sum_{i=1}^{|V|} y_t^i \log \frac{e^{w_t^i}}{Z_t} = \frac{1}{T} \sum_{t=1}^T \sum_{i=1}^{|V|} y_t^i (\log{Z_t} - {w_t^i}). 
\end{equation}

During training, we assume that $y$ is a one-hot distribution since we do not have access to the true data distribution.  This assumption is obviously incorrect, considering that humans are able to form a valid continuation of most contexts with multiple potential words.  Call these potential words the {\em branching set} of a given context.  Instead, we may want to give partial credit 
for predicting words in the branching set, like `{\em rug}' in instances when the GT is `{\em mat}' (and vice versa) based on the semantic similarity between the words and contexts. One method for achieving this is via {\em label smoothing}.

\subsection{Label smoothing and knowledge distillation}\label{sec::LS-KD}

Label smoothing (LS) is a method which modifies each $y_t$ to be soft-targets, instead of one-hot, by redistributing probability from the GT label to other labels \citep{szegedy2016rethinking}.  It is commonly implemented using a hand-crafted function, such as setting the GT probability to 0.9 and spreading the remaining 0.1 uniformly across the other labels \citep{pereyra2017regularizing}.  LS is a simple method which has different motivations depending on the problem it is meant to solve.  One problem is noisy labels \citep{pmlr-v119-lukasik20a}.  Here one views the GTs as truly one-hot but possibly incorrect due to errors in annotation.  A second problem is to regularize over-confident models in situations where we desire that model's probabilities should not spike to a single prediction \citep{pereyra2017regularizing, muller2019does}.  Here, one again views the GTs as truly one-hot but they do not want the model to make predictions too close to the one-hot distribution due to how the model is to be used.  However, in our case, we use label smoothing to make the single-target GTs multi-target, under the view that the true GTs are actually multi-target but that we lack access to this information.

Given that the multiple-targets should change given the context, we desire the smoothing values to depend on the context.  One can have a data-dependent or semantically-aware LS by using an auxiliary pre-trained model to form label scores, such as using cosine scores from FastText \citep{elbayad2018token, li2019data}.  This can be seen as a form of knowledge distillation (KD), which is when a pre-trained teacher model is used to train a student model by minimizing the cross entropy, $H(p_{\mathrm{KD}}, p)$, between the teacher's distribution, $p_{\mathrm{KD}}$ and the student's, $p$ \citep{hinton2015distilling, wang2020knowledge}.  Here, the smoothing scores are derived from the teacher's distribution and thus can be viewed as parametric form of semantic LS \citep{yuan2020revisiting, tang2020understanding}.  


One can perform KD either from a weaker language representation model such as FastText or a strong language model such as GPT-2.   However, this raises the question -- \emph{who teaches the teacher}?  That is, in order to train our LM we need to assume the existence of an already established model, which seems to defeat the purpose.  Instead, we set our own desideratum that we do not want to use a pre-trained neural LM as a teacher.


We hypothesize that $N$-gram statistics are a rich source of semantic information which can be exploited 
to train neural LMs, i.e., introducing global $N$-gram statistics to our otherwise local training procedure \citep{neubig2016generalizing, zhao-etal-2017-ngram2vec, yang2019using}.  The na\"ive approach would be to pre-train an $N$-gram LM to use as our teacher; however, this will not work because the $N$-gram model will be a weak teacher.  $N$-gram LMs perform worse than neural language models \citep{FITPT283, jozefowicz2016exploring}.  Because of this, if we just try to match the teacher's distribution, as is done in KD, we will end up with a higher perplexity model than had we forgone the KD from the $N$-gram model.  To overcome this issue, we need to examine how label smoothing and $N$-gram LMs work.

\subsection{Ranking in lieu of label smoothing}

We can break semantic label smoothing into two problems: a) how to choose which words should get partial credit in a given context, and b) how much partial credit they should receive. To solve the first problem, we could use the common `distributional hypothesis' from linguistics to give partial credit to words which share similar contexts  \citep{harris1954distributional, mikolov2013distributed}.  
That is, words which have some shared context should get some partial credit.  The distributional hypothesis underlies $N$-gram language models as well, which employ smoothing algorithms  and back-off methods to weigh higher-order $N$-grams more than lower order ones \citep{goodman2001bit}. To solve the problem of quantifying credit assignment, it seems we need a probabilistic (even if un-normalized) model to specify the smoothing values.  However, as reasoned above, if we na\"ively use $N$-gram LMs for this, we will wind up trying to match the distribution of a weaker teacher model.  One solution to this would be to modify the $N$-gram LM teacher itself. 

$N$-gram LM algorithms were developed with the goal of the model being a valid LM for use in applications.  One criterion of such a model is that it produces a non-zero probability\footnote{Though this may not be true for current applications of LMs.}.  This is achieved by including the unigram distribution.  Another criterion is that the $N$-gram LM generalizes well to unseen contexts.  This is achieved by well-developed smoothing algorithms.  If we only intend to use the $N$-gram model as a teacher, we may ease or remove these criteria.  For example, we might posit the converse of the distributional hypothesis and say that we will restrict credit from words which do not share contexts.  To achieve this, we could forego backing-off to the unigram distribution and smoothing into unseen contexts. One of the main motivations of neural LMs over $N$-grams is their ability to generalize due their distributed representations \citep{bengio2003neural}. This implies that we could forego applying any kind of smoothing algorithm to the $N$-gram teacher and let the neural student learn to generalize to unseen contexts.  If we follow this approach, then we would need to make modifications to existing $N$-gram LM algorithms to determine  how to combine the various $N$-gram probabilities into a (sparse) probabilistic model.  Instead, we propose a more radical approach in order to avoid this.


If we decompose the problem of specifying label weights, we are specifying not only how to weight each cost term but also, implicitly, an ordering of the weighted terms.  Thus, given label  weights, we can convert the problem of LS to a ranking problem by specifying that the most probable label is the most relevant rank and the second most probable is the second most relevant rank, etc. 
{\bf The main contribution of this paper is therefore the idea that we may not need to specify specific label weights, so long as we can specify an {\em order} to the labels.}  In particular, we hypothesize that the ordering of labels contains sufficient semantic information to be used for our desired partial credit. To specify the ordering, we again employ the distributional hypothesis and consider that increasing the shared context between a word and the ground-truth word increases the relevance of that word in the ordering.  Note that the original GT word will always retain the most relevant rank, as it will always share the most context with itself.  Since ordinal information is less strict than interval information, we may overcome the issue of having a weak teacher since we are no longer trying to directly match a weaker distribution.  

This approach obviates the need for a modified probabilistic $N$-gram LM since we just need to consider the size of a shared $N$-gram context to determine how much credit to give to a set of words which can follow.   We describe how to construct artificial ranking GTs from $N$-gram statistics in Section \ref{sec:bsconst} and how to use these in training the neural LM in Section \ref{sec::pl-loss}.

\begin{figure*}[!htbp]
\centering
\small
\begin{center}
\begin{tabular}{l|l|l|l|l|l|l|l|l|l} 
Adams & River & is & a & tributary & to & the & Thompson & and & Fraser \\ \hline

\cellcolor[rgb]{0.651,0.808,0.890}{Moro} & \cellcolor[rgb]{0.698,0.875,0.541}{passes} & \cellcolor[rgb]{0.698,0.875,0.541}{Lumber} & \cellcolor[rgb]{0.698,0.875,0.541}{the} &  & \cellcolor[rgb]{0.651,0.808,0.890}{,} & \cellcolor[rgb]{0.200,0.627,0.173}{Jim} &  & \cellcolor[rgb]{0.698,0.875,0.541}{sub} & \cellcolor[rgb]{0.698,0.875,0.541}{Sam} \\ \arrayrulecolor{white}\hline 

\cellcolor[rgb]{0.651,0.808,0.890}{Shebelle} & \cellcolor[rgb]{0.200,0.627,0.173}{Lake}  & \cellcolor[rgb]{0.698,0.875,0.541}{$<$unk$>$} & \cellcolor[rgb]{0.698,0.875,0.541}{located} &  & \cellcolor[rgb]{0.651,0.808,0.890}{of} & \cellcolor[rgb]{0.200,0.627,0.173}{Harry} &  &  & \cellcolor[rgb]{0.200,0.627,0.173}{$<$unk$>$}  \\ \arrayrulecolor{white}\hline 

\cellcolor[rgb]{0.651,0.808,0.890}{Colorado} &  & \cellcolor[rgb]{0.698,0.875,0.541}{to} & \cellcolor[rgb]{0.698,0.875,0.541}{thought} &  & \cellcolor[rgb]{0.122,0.471,0.706}{the} &  &  &  & \cellcolor[rgb]{0.200,0.627,0.173}{Kansas} \\ \arrayrulecolor{white}\hline

\cellcolor[rgb]{0.651,0.808,0.890}{Missouri} &  & \cellcolor[rgb]{0.698,0.875,0.541}{run} & \cellcolor[rgb]{0.698,0.875,0.541}{north} &  & \cellcolor[rgb]{0.122,0.471,0.706}{in} &  &  &  & \cellcolor[rgb]{0.200,0.627,0.173}{Platte}\\ \arrayrulecolor{white}\hline 

\cellcolor[rgb]{0.200,0.627,0.173}{Payette} &  & \cellcolor[rgb]{0.698,0.875,0.541}{flows} & \cellcolor[rgb]{0.698,0.875,0.541}{298} &  & \cellcolor[rgb]{0.122,0.471,0.706}{valleys} &  &  &  & \cellcolor[rgb]{0.200,0.627,0.173}{Missouri}\\ \arrayrulecolor{white}\hline

\cellcolor[rgb]{0.200,0.627,0.173}{$<$unk$>$} &  & \cellcolor[rgb]{0.698,0.875,0.541}{begins} &  &  & \cellcolor[rgb]{0.122,0.471,0.706}{took} &  &  &  & \cellcolor[rgb]{0.200,0.627,0.173}{Yellowstone} \\ \arrayrulecolor{white}\hline

\cellcolor[rgb]{0.200,0.627,0.173}{Back} &  & \cellcolor[rgb]{0.698,0.875,0.541}{(} &  &  & \cellcolor[rgb]{0.122,0.471,0.706}{by} &  &  &  & \cellcolor[rgb]{0.200,0.627,0.173}{Mickey} \\ \arrayrulecolor{white}\hline

\cellcolor[rgb]{0.200,0.627,0.173}{Red} &  & \cellcolor[rgb]{0.698,0.875,0.541}{valley} &  &  & \cellcolor[rgb]{0.122,0.471,0.706}{paymen-} &  &  &  & \cellcolor[rgb]{0.200,0.627,0.173}{Mississippi} \\ \arrayrulecolor{white}\hline 

\cellcolor[rgb]{0.200,0.627,0.173}{Yellowst-} &  & \cellcolor[rgb]{0.698,0.875,0.541}{Provincial} &  &  & \cellcolor[rgb]{0.122,0.471,0.706}{is} &  &  &  & \cellcolor[rgb]{0.200,0.627,0.173}{Cheyenne} \\ \arrayrulecolor{white}\hline

\cellcolor[rgb]{0.200,0.627,0.173}{Grand} &  & \cellcolor[rgb]{0.698,0.875,0.541}{sockeye} &  &  &  &  &  &  & \cellcolor[rgb]{0.200,0.627,0.173}{Columbia} \\ \arrayrulecolor{white}\hline

\cellcolor[rgb]{0.200,0.627,0.173}{Mississ-} &  & \cellcolor[rgb]{0.698,0.875,0.541}{area} &  &  &  &  &  &  & \cellcolor[rgb]{0.200,0.627,0.173}{Ohio} \\ \arrayrulecolor{white}\hline 

 &  & \cellcolor[rgb]{0.200,0.627,0.173}{,} &  &  &  &  &  &  & \cellcolor[rgb]{0.200,0.627,0.173}{Osage} \\ \arrayrulecolor{white}\hline
 
 &  & \cellcolor[rgb]{0.200,0.627,0.173}{and} &  &  &  &  &  &  & \cellcolor[rgb]{0.200,0.627,0.173}{Kissimmee} 
\end{tabular}
\end{center}
\caption{Branching sets created for the ground-truth `{\em Adams River is a tributary to the Thompson and Fraser Rivers ...}' using the ground-truths ($O_{gt}$) and four orders ({\setlength{\fboxsep}{0pt}\colorbox[rgb]{0.651,0.808,0.890}{$O_5$}}, {\setlength{\fboxsep}{0pt}\colorbox[rgb]{0.698,0.875,0.541}{$O_4$}}, {\setlength{\fboxsep}{0pt}\colorbox[rgb]{0.200,0.627,0.173}{$O_3$}}, {\setlength{\fboxsep}{0pt}\colorbox[rgb]{0.122,0.471,0.706}{$O_2$}})  from  the Wiki02 training partition.  
The branching sets for `{\em tributary}' and `{\em Thompson}' are only the ground-truths since all other orders have been pruned. 
} 
\label{fig::wiki-ngram-sample-1-bi400} 
\end{figure*}

\section{Methods}\label{Methods}

\subsection{$N$-gram branching set construction} \label{sec:bsconst}

Let a {\em branching set}  (BS), $b_t$, be a set of words which can continue a given context $c_t=(x_1, \dots, x_{t-1})$ at step $t$.  
Let $c_t(n)$ be a slice of the context which only includes the previous $n-1$ words (assuming $t \geq n$).  Intuitively, the construction of the BS is simple.  We are going to consider a series of past $N$-gram contexts, starting with the largest context until the smallest, and any word that we find in the training set that continues those contexts will be in the BS.  Then, in order to derive artificial rank GTs from the BS, we can  give a \emph{weak} ordering\footnote{For terminology, we use  `full ordering' to indicate known ranks for all $|V|$ words and `partial ordering' to indicate known top-$k$ ranks.  A `strong ordering' indicates there are no ties and a `weak ordering' indicates that there are either ties or an unknown preference across contiguous ranks. This is also known as a `partitioned preference'.} to the words in the BS by specifying that words in the BS with a larger context should have a higher rank relevance than those with a smaller context. If, for example, we consider bigrams and trigrams then any word that continues the trigram context will also continue the bigram context but words which only continue the bigram context (but not the trigram context) will be less relevant.  Unigrams are excluded since they use no context.

More formally, we construct a tuple of sets referred to as \emph{orders}, $O = (O_{gt}, O_M, \dots, O_{2})$ which will be merged to form the BS.  $O_{gt} = \{x_{t}\}$, only contains the original GT.
The other orders are made up of all words which continue a specific context length, 
such that we construct each of the orders as $O_m = \{x | x \in V \setminus O_{>m}, C(c_t(m), x) \geq 1\}$,
where $m \in \{M, \dots, 2\}$ and $C(c_t(m), x)$ is the count for how many times $x$ appears in the context $c_t(m)$.   Note, we exclude words already included in previous orders.  
We also use a cut-off, $q$, so that we only consider sufficiently small orderings  $|O_m| < q$, since large BSs will not be informative; e.g., all words which follow `{\em the}' or `{\em is a}'.  
The BS is constructed as an ordered tuple where $b_t = (x \,|\, x \in O_i, O_i \in O )$. 
The artificial GT ranks are then specified as the weak rank ordering of the BS, which is weak since each $O_m$ may contain multiple unordered words.


 We can further resolve the rank within each order by using future context in addition to past context.  This assumes that, had we 
 access to longer past contexts, the ranks would have resolved the same way as using future information.  Note that we can introduce future information here because the teacher model does not need to form a valid probability distribution since it only defines targets for the student model.  Let $c_t(n_p, n_f)$ be the context defined by using $n_p-1$ words in the past and $n_f-1$ words in the future.  We prioritize past context over future context, since that is what the student model will be able to observe as input.
This is done by specifying the orders by a reverse orthographical sorting of past and future context lengths, i.e., $c_t(3, 4) \succ c_t(3, 3) \succ c_t(2, 4)$. 
This does not completely resolve the weak ordering, though, since two words may share both the same past and future contexts.
 Figure \ref{fig::wiki-ngram-sample-1-bi400} shows an example BS and Figure \ref{fig::wiki-ngram-sample-2-bi400} shows how it was constructed using future information.   We present a memory-efficient algorithm for creating the rank GTs in Appendix \ref{app::mem}.



\begin{figure*}[htbp]
\centering
\begin{center}
\begin{tabular}{lll|ll:l:l||ll:l:l|} 
Adams & River & is & a & tributary & to & the & Thompson & and & Fraser & Rivers \\ \hline
 &  &  & \cellcolor[rgb]{0.651,0.808,0.890}{a} & \cellcolor[rgb]{0.651,0.808,0.890}{tributary} & \cellcolor[rgb]{0.651,0.808,0.890}{,} & \cellcolor[rgb]{0.651,0.808,0.890}{the} & \cellcolor[rgb]{0.698,0.875,0.541}{Thompson} & \cellcolor[rgb]{0.698,0.875,0.541}{and} & \cellcolor[rgb]{0.698,0.875,0.541}{Sam} &  \\ \arrayrulecolor{white}\hline 
 
 &  &  & \cellcolor[rgb]{0.651,0.808,0.890}{a} & \cellcolor[rgb]{0.651,0.808,0.890}{tributary} & \cellcolor[rgb]{0.651,0.808,0.890}{of} & \cellcolor[rgb]{0.651,0.808,0.890}{the} &  & \cellcolor[rgb]{0.200,0.627,0.173}{and} & \cellcolor[rgb]{0.200,0.627,0.173}{$<$unk$>$} & \cellcolor[rgb]{0.200,0.627,0.173}{Rivers} \\ \arrayrulecolor{white}\hline 
 
 &  &  &  & \cellcolor[rgb]{0.122,0.471,0.706}{tributary} & \cellcolor[rgb]{0.122,0.471,0.706}{the} &  &  & \cellcolor[rgb]{0.200,0.627,0.173}{and} & \cellcolor[rgb]{0.200,0.627,0.173}{Kansas} & \cellcolor[rgb]{0.200,0.627,0.173}{Rivers} \\ \arrayrulecolor{white}\hline
 
 &  &  &  & \cellcolor[rgb]{0.122,0.471,0.706}{tributary} & \cellcolor[rgb]{0.122,0.471,0.706}{in} &  &  & \cellcolor[rgb]{0.200,0.627,0.173}{and} & \cellcolor[rgb]{0.200,0.627,0.173}{Platte} & \cellcolor[rgb]{0.200,0.627,0.173}{Rivers} \\ \arrayrulecolor{white}\hline
 
 &  &  &  & \cellcolor[rgb]{0.122,0.471,0.706}{tributary} & \cellcolor[rgb]{0.122,0.471,0.706}{valleys} &  &  & \cellcolor[rgb]{0.200,0.627,0.173}{and} & \cellcolor[rgb]{0.200,0.627,0.173}{Missouri} & \cellcolor[rgb]{0.200,0.627,0.173}{Rivers} \\ \arrayrulecolor{white}\hline
 
 &  &  &  & \cellcolor[rgb]{0.122,0.471,0.706}{tributary} & \cellcolor[rgb]{0.122,0.471,0.706}{took} &  &  & \cellcolor[rgb]{0.200,0.627,0.173}{and} & \cellcolor[rgb]{0.200,0.627,0.173}{Yellowstone} & \cellcolor[rgb]{0.200,0.627,0.173}{Rivers} 
 
 
 
\end{tabular}
\end{center}
\caption{BS construction for the example in Figure \ref{fig::wiki-ngram-sample-1-bi400} using four orders, {\setlength{\fboxsep}{0pt}\colorbox[rgb]{0.651,0.808,0.890}{2p-1f}}, {\setlength{\fboxsep}{0pt}\colorbox[rgb]{0.698,0.875,0.541}{2p}}, {\setlength{\fboxsep}{0pt}\colorbox[rgb]{0.200,0.627,0.173}{1p-1f}}, {\setlength{\fboxsep}{0pt}\colorbox[rgb]{0.122,0.471,0.706}{1p}} and the GTs.  Where `{\setlength{\fboxsep}{0pt}\colorbox[rgb]{0.651,0.808,0.890}{2p-1f}}' indicates an order which matches two past words and one future word. We show the branching sets for `to' and `Fraser', centred in dash lines, and the context that selected those words in solid lines.  The table has been truncated. 
} 
\label{fig::wiki-ngram-sample-2-bi400} 
\end{figure*}

\subsection{Plackett-Luce rank loss}\label{sec::pl-loss}

Let us assume that the BS, $b_t$, is strongly ordered by some preferred (partial) rank over $k \leq |V|$ words in $V$ for step $t$ and that the GT word has the first rank.  Let $y_t$ be the set of words in $V$ and say that $b_t(m)$ indexes the element of rank $m$ according to $b_t$, i.e, $b_t$ defines a rank permutation.  If we assume that $y_t$ and $w_t$ are sorted by rank, we can drop the indexing in the following section.  


Given strongly ordered top-$k$ ground-truths,  we train our LM by learning to rank with the top-$k$ Plackett-Luce (PL) rank loss  \citep{plackett1975analysis, cao2007learning, xia2009statistical}. The PL loss models the data as a Plackett-Luce distribution, where for any time step $t$,

\begin{equation}\label{eq::PLD}
\begin{split}
p_t(y_t^{1} \succ \dots \succ y_t^{k}) & = \prod_{i=1}^{k} p_t(y_t^{i} \,|\, y_t^{1}, \dots, y_t^{i-1} )  = \prod_{i = 1}^{k} \frac{e^{w_t^{i}}}{\sum_{j = i}^{|V|} e^{w_t^{j}}}.
\end{split}
\end{equation}
The PL distribution  represents a generative process of sampling from a softmax distribution without replacement, were we re-normalize the distribution by excluding previous samples at every ranking step.  
The loss is then defined as the negative log-likelihood of equation \ref{eq::PLD},




\begin{equation}
-\log p_t(y_t^1 \succ  \dots \succ y_t^{k})= \sum_{i = 1}^{k} \log \sum_{j=i}^{|V|} e^{w_t^{j}} - {w_t^{i}} = \sum_{i = 1}^{k} \log\Big(Z_t - \sum_{j < i} e^{w_t^{j}}\Big) - {w_t^{i}}.
\end{equation}

While there are many surrogate ranking losses, we choose PL since, in the case when $k=1$, the PL distribution collapses to the softmax distribution and the loss to the corresponding CE loss.  That is, we revert back to traditional language modelling.   This is a beneficial property since a) we might encounter contexts during training without available ranking data, i.e., only the original GT word is in the branching set, b) we only consider the probability of the original GT during evaluation, 
and c) we do not wish to radically change the distribution being modelled so that PPL is comparable to 
previous work.  The PL loss can also be efficiently implemented for GPUs and does not add significant computation over a regular softmax (see Appendix \ref{app::pl-alg}).

When combined with a method for constructing artificial ranking ground-truths, we can view the PL loss as a form of self-supervision where the first ranking term is our original loss and all higher ranking terms are auxiliary losses.  However, unlike most auxiliary losses used in self-supervised learning, the extra PL loss terms are a natural extension or generalization of the original loss.

The PL loss can fail to capture the inductive bias that it is more important to get the more-preferred ranks correct over the lesser-preferred ranks
\citep{lan2014position}. This is  relevant since the original GT is the first rank and we will have less confidence as we descend the ranks, since the amount of semantic information in a bigram is generally less than in a trigram, and so on.  \citet{lan2014position} introduced discount weights on the PL terms which decreased exponentially, but,
since their exact method did not work for our task, we use a {\em stepped} function
where a static $\eta < 1$ weight is given to the top rank 
and the remaining $1 - \eta$ is partitioned such that there is an equal difference between the weights for consecutive ranks. 
In practice, $\eta$ acts similar to the temperature smoothing hyper-parameter used in KD.




The PL loss does not allow for weak orderings.  This is a problem since the words within each individual order, $O_i$, have no ordering; i.e., if $O_i = \{o_1, \dots, o_S\}$, then the $S$ items are of an unknown ordering.  This creates a partitioned preference where all the partitions define a ranking, $O_i \succ O_{i+1}$, but the ranking within each ordering is unknown. One way of handling this would be to consider the marginal distribution across all permutations for each ordering.  However, the likelihood would require calculating a factorial number of terms \citep{ma2020learning}. 
Instead, we consider a lack of known preference as an inductive bias that words in the weak orders are equally preferred.  We enforce this by modifying the PL loss such that we revert back to regular softmax-cross entropy within the weak order for rankings $i, \dots, i+S$ as $ \sum_{s = i}^{i + S} \log Z_{t,i} - {w_t^{s}}$, where $Z_{t,i} = Z_t - \sum_{j<i} e^{w_t^{j}}$, which will optimize to a uniform distribution within the ordering.  This is a novel modification to our knowledge.

\section{Related work}


Previous work combined $N$-gram information with neural LMs.  
\citet{neubig2016generalizing, bakhtin2018lightweight} used dynamic mixtures of $N$-grams and neural LMs. \citet{neubig2016generalizing} found that their models performed worse than regularly trained neural LMs except on low-frequency words.  \citet{noraset2018controlling} used soft constraints for text generation with an LSTM LM.  They tried two sets of constraints: those to reduce repeated words during generation and those to match Kneser-Ney bigram probabilities during generation.  Their method  estimates the marginal probability of the LSTM LM given some conditioning constraint and regularizes it using a KL-divergence against the constraint distribution.  These marginal probabilities are based on a pool of generated text that is periodically updated during training.  This can be seen as a form of LS against the actual generative distribution instead of the GT distribution.  They showed they could better match the distribution of repeated tokens and bigram statistics against a baseline on PTB.  
\citet{yang2019using} built on this 
by making computation of the marginals more tractable. 
They regularized the LM's whole vocabulary against trigram statistics, which makes their methods very similar to KD against a trigram $N$-gram model.  
They also proposed a method for selecting useful $N$-grams based on an information criterion.  This is important, as they need to do a single forward pass for every $N$-gram context they regularize.  So selecting which contexts they use significantly decreases the computational cost.  They trained an LSTM LM and a trigram $N$-gram LM.  On Wiki02, their method achieved 69 PPL given a baseline of 76.

Various works implemented semantic label smoothing using pre-trained probabilistic models \citep{elbayad2018token, hahn-choi-2019-self, li2019data, ghoshal2020learning, liu2020noisy, lukasik-etal-2020-semantic}.  These can use various pre-trained models such as embedding models like Fasttext \citep{joulin2017bag}, language representation models like BERT \citep{devlin2019bert} or language models like GPT2 \citep{radford2019language}. 
\citet{lukasik-etal-2020-semantic} smoothed over related sequences which were found using a pre-trained model and then refined using BLEU as a selection criterion.  \citet{wang-etal-2020-inference} introduced graduated label smoothing which uses a set of binned label-smoothing values and the confidence of a pre-trained model to determine the bin to which each token belongs,  improving BLEU and calibration for NMT.    

\citet{li2019data} used an auxiliary KL-divergence based loss on a semantically-aware or data-dependent 
Gaussian prior which is parameterized using cosine similarity from Fasttext embeddings.  They applied this over a range of conditional language modelling tasks including NMT, text summarization, and story telling.  For story telling, their method lowered test PPL by $\approx2$-$3$  points.  In general, LS can be framed as an entropy regularizer and many prior distributions can be used \citep{pereyra2017regularizing, meister-etal-2020-generalized}. 

\citet{ghoshal2020learning}  proposed a method which  jointly trains a model with a set of learned smoothing  parameters.  Given $C$ classes,  one could implement semantic LS by learning a similarity $C \times C$ matrix where each row provides smoothing scores for a given target class.  Instead, the authors approximated this with a $C \times k$ matrix with  $k \lll C$.  While this is a form of semantic label smoothing, 
it is limited in that it is based only on the target class and not the full context.  They applied it to semantic parsing and question answering.  

\citet{welleck2019neural} introduced an unlikelihood loss which penalizes words which should have low probability.  We make use of the concept of a branching set which is the words that can follow a given context.  The complementary set of the branching set could be used to select negative samples for the unlikelihood loss.   

Born-again distillation or self-distillation is where the student and teacher have the same specification \citep{furlanello2018born, hahn-choi-2019-self, yuan2020revisiting}.  \citet{furlanello2018born} applied born-again KD to LMs and showed that a student LSTM LM on PTB could outperform the same teacher model. \citet{yang2019training} used a top-$k$ auxiliary loss and trained image classification models with a series of self-distillations.  \citet{tang2020understanding} developed a top-$k$ KD using the teacher's probability for the top-$k$ ranks and distributing the remaining probability uniformly.

\citet{pmlr-v130-reddi21a} recently introduced the use of top-$k$ rank loss functions for rank-based KD using PL and pairwise hinge losses,   outperforming traditional KD on a variety of  traditional ranking tasks.  
Other forms of structured KD can also be applied to other NLP sequence tasks as well as computer vision \citep{wang-etal-2021-structural, DBLP:conf/cvpr/LiuCLQLW19}.  \citet{cole2021multi} considered multi-label image classification when only having access to singular-class GTs and thus has a similar motivation to our work.  




\section{Experiments}

\begin{table*}[htbp]
\centering
\begin{tabular}{llrlllll}
Data & Model & (P-)PPL & $\mathrm{A }\subseteq 1$ & $\mathrm{A }\subseteq 2$ & $\mathrm{A }\subseteq 3$ & $\mathrm{A }\subseteq 5$ & $\mathrm{A }\subseteq 10$ \\

\hline \multirow{3}{*}{PTB}
& BERT 24-layer& *1.205& 0.975& 0.986& 0.989& 0.991& 0.993\\
& GTP-2 774M& 22.810& 0.437& 0.525& 0.573& 0.630& 0.701\\
& BA-LSTM & 60.724  & 0.302 & 0.394 & 0.446 & 0.511 & 0.593 \\
\hline \multirow{4}{*}{Wiki02}
& BERT 24-layer& *1.423&  0.930& 0.979& 0.983& 0.985& 0.987\\
& GTP-2 774M& 25.277& 0.419& 0.521& 0.575& 0.635& 0.706\\
& BA-LSTM & 68.437  & 0.296 & 0.394 & 0.448 & 0.512 & 0.592\\
& BA-T-XL & 67.468  & 0.301 & 0.398 & 0.452 & 0.517 & 0.598\\



\end{tabular}
\caption{Perplexity (PPL) and accuracy in the top-$k$ ($\mathrm{A }\subseteq k$ ) for word-level fined-tuned probabilistic teacher models BERT, GPT-2, Born-Again (BA) 
on the PTB and Wiki02 validation sets.  *We report a {\em pseudo}-perplexity (P-PPL) for BERT as it does not form a valid distribution over sequences.}  
\label{tab::bertgpt2-per-valid}
\end{table*}




We use the word-level Penn Treebank (PTB) and WikiText-02 (Wiki02) datasets and 
use ADW-LSTM (LSTM) and Transformer-XL (T-XL) students \citep{taylor2003penn, DBLP:conf/iclr/MerityX0S17, DBLP:journals/corr/abs-1803-08240, dai-etal-2019-transformer}.  
Our LSTM uses 24.2M parameters on PTB and 33.5M on Wiki02, and our  
Transformer-XL uses 35.8M on Wiki02.  

We propose two hypotheses: that we can re-frame the problem of label-smoothing as a form of rank-based knowledge distillation from a teacher which {\em only} provides rank ground-truths and that we can derive the ranks directly from a non-probabilistic $N$-gram teacher.  In order to decouple these two hypotheses, we first use three different types of pre-trained LMs to evaluate the PL loss when used for KD then, provided that works, we apply the PL loss with an $N$-gram teacher model.

We choose GPT-2, BERT, and Born Again (BA) models as teachers for different reasons.  GPT-2 and BERT were chosen under the assumption that these large LMs will produce better ranks than the $N$-grams.  We tried both since the former is an auto-regressive LM which only conditions on past context and thus matches our student models, while the latter is an auto-encoding language representation model using both past and future contexts,   allowing us to test if we can distil future information. 
BA models are also considered as they will not present data-leakage problems, unlike BERT and GPT-2 which were pre-trained on a large amount of auxiliary data.  The BA models are selected as the highest performing CE baseline models.

GPT-2 and BERT use sub-word vocabularies instead of the standard word-level ones for PTB and Wiki02.  We convert them  by summing the sub-word hidden states to get whole-words and then fine-tuning them using CE. 

Table \ref{tab::bertgpt2-per-valid}  shows the validation performance of the teacher models.
The performance of these models is better than the baseline CE models, which  warrants their use as teacher models.  
BERT out-performs GPT-2 due to using surrounding context when predicting a word where GPT-2 is limited to only past context\footnote{Technically, these PPL results are not directly comparable since BERT does not form a valid distribution over sequences, hence using P-PPL for BERT.  Also, they use different auxiliary training data.}.   We provide two other sanity checks; first, we provide examples of the top-$k$ predictions for BERT, GPT-2 and the $N$-grams  in Appendix \ref{app::topk-samples} and, second, we plot top-$k$ frequency statistics for BERT and GPT-2 in Appendix \ref{app::freq-dist}.

The $N$-grams used a set of contexts from 5 to 1 past tokens concurrently with 4 to 0 future tokens and a pruning cut-off of 10, i.e. 5p-4f, 5p-3f, $\dots$, 1p-1f, 1p.

We compare a CE baseline to four other losses.  The first is a top-$k$ KL-based KD.  This has been modified in three ways from traditional KD. First, we only use the top-$k$ ranks, as did \citet{tang2020understanding}, although we forgo placing a uniform distribution over all words not in the top-$k$.  They reported a test PPL of 60.85 using top-$k$ KD on PTB with AWD-LSTM using a smaller student with 9.1M parameters. 
Second, we post-process the predicted ranks by floating the original GTs to the top rank while keeping the top-$k$ probabilities the same order i.e. make the GTs the most probable. 
Initial experimentation did not show a significant change for KL-based KD and we believed this was an important modification for rank-based KD.  Third, we cycle the interpolation value between the CE and KD terms, similar to \citet{clark-etal-2019-bam, jafari-etal-2021-annealing}.  
For PL, we can forego discounting (PL), use the teacher's probabilities (PL-t), or use the stepped function (PL-s).  Note that the KL also uses the teacher's probabilities. See Appendix \ref{app::imp-details} for further experimental details.

\begin{table*}[htbp]
\footnotesize
\begin{center}
\resizebox{\textwidth}{!}{\begin{tabular}{ll|ll|ll|ll|ll|ll|ll}

& &  \multicolumn{4}{c|}{\bf{PTB}} &  \multicolumn{8}{c}{\bf{Wiki02}}\\

& &  \multicolumn{4}{c|}{LSTM Student} &  \multicolumn{4}{c|}{LSTM Student} & \multicolumn{4}{c}{ T-XL Student}\\
 & &  \multicolumn{2}{c|}{Validation} & \multicolumn{2}{c|}{Test} &  \multicolumn{2}{c|}{Validation} & \multicolumn{2}{c|}{Test} &  \multicolumn{2}{c|}{Validation} & \multicolumn{2}{c}{Test}\\ 
T & Loss & PPL & SEM & PPL & SEM & PPL & SEM & PPL & SEM & PPL & SEM & PPL & SEM\\ \hline
\multirow{1}{*}{} & CE &  61.44  & 0.014 &  59.11  & 0.015 &  69.59  & 0.018 &  66.43  & 0.015 &  68.66  & 0.021 &  65.27  & 0.020 \\

\hline
 
\multirow{4}{*}{\rotatebox[origin=c]{90}{GPT-2}} & KL & \underline{57.43}  & 0.012 & \underline{55.46}  & 0.010 & \underline{65.41}  & 0.008 & \underline{62.79}  & 0.007 & \underline{59.43}  & 0.009 & \underline{56.70}  & 0.008 \\

 & PL-t  & \underline{57.28}  & 0.014 & \underline{55.17}  & 0.013 & \underline{65.47}  & 0.015 & \underline{62.69}  & 0.014 & \underline{59.32}  & 0.007 & \underline{56.58}  & 0.007 \\
 
 & PL & \underline{58.16}  & 0.012 & \underline{56.25}  & 0.012 & \underline{65.45}  & 0.011 & \underline{62.74}  & 0.011 & \bf \underline{58.61}  & 0.007 & \bf \underline{55.94}  & 0.007 \\
 
 & PL-s & \underline{57.63}  & 0.008 & \underline{55.67}  & 0.007 & \underline{65.34}  & 0.008 & \bf \underline{62.59}  & 0.008 & \underline{59.48}  & 0.008 & \underline{56.76}  & 0.007 \\
 \hline
 
\multirow{4}{*}{\rotatebox[origin=c]{90}{BERT}} & KL &  59.86  & 0.018 &  57.73  & 0.017 &  68.63  & 0.010 &  65.51  & 0.009 &  67.79  & 0.010 &  64.16  & 0.009 \\

 & PL-t & \bf \underline{59.38}  & 0.009 & \bf  57.20  & 0.009 & \bf  68.04  & 0.010 & \bf  64.99  & 0.009 &  67.66  & 0.012 &  64.18  & 0.012 \\
 
 & PL &  62.23  & 0.011 &  60.32  & 0.011 &  71.98  & 0.009 &  68.68  & 0.008  &  67.78  & 0.009 &  64.28  & 0.010 \\
 
 & PL-s &  61.13  & 0.010 &  59.11  & 0.010 &  68.55  & 0.007 &  65.53  & 0.008 &  67.72  & 0.008 &  64.26  & 0.008 \\
 \hline
 
\multirow{4}{*}{\rotatebox[origin=c]{90}{BA}} & KL & \bf  60.14  & 0.011 & \bf  57.63  & 0.010  &  68.90  & 0.014 &  65.75  & 0.012 &  67.46  & 0.014 &  64.19  & 0.012 \\

 & PL-t &  60.80  & 0.014 &  58.15  & 0.014 &  69.14  & 0.009 &  65.74  & 0.009 & \bf  66.96  & 0.012 & \bf 63.46  & 0.011 \\
 
 & PL &  63.36  & 0.017 &  60.54  & 0.013 &  69.63  & 0.007 &  66.16  & 0.008 & \bf  66.93  & 0.017 & \bf  63.58  & 0.014 \\
 
 & PL-s &  60.91  & 0.010 &  58.22  & 0.012 & \bf  67.96  & 0.007 & \bf  64.76  & 0.006 & \bf  67.10  & 0.012 & \bf  63.64  & 0.011 \\
 \hline
 
\multirow{3}{*}{\rotatebox[origin=c]{90}{$N$-gram}} & PL-s &  59.87  & 0.022 &  57.33  & 0.021 &  67.58  & 0.008 &  64.82  & 0.007 &  66.86  & 0.017 &  63.87  & 0.016 \\
& wPL-s & 59.66  & 0.008 &  57.16  & 0.009 & 67.25  & 0.006 &  64.44  & 0.006 &  66.59  & 0.015 &  63.54  & 0.013 \\
& wPL & & & & & & & & &  66.38  & 0.013 &  63.49  & 0.012 \\
 \hline
 
\end{tabular}}
\caption{ Average PPL ($\downarrow$) with standard error (SEM) ($n=30$) using an LSTM and T-XL student models and GPT-2, BERT, Born again and N-gram teachers (T).  
wPL is our version of Plackett-Luce modified to account for weak orderings.
Bolded PL experiments exceed the in-group KL baseline with $p < .001$ using a two-tailed t-test.  Bolded KL experiments exceed the best in-group PL experiment with significance.  
Underlined experiments are those which perform better than the $N$-gram model with significance.  We skipped two of the wPL experiments since only the T-XL student seems to benefit from no discounting.  
}
\label{tab::Wiki02-table}
\end{center}
\end{table*}

\section{Discussion}

The results in Table \ref{tab::Wiki02-table} show we can treat language modelling as a structured prediction task using a rank-based KD and that this can be done without the need of a probabilistic teacher.  We discuss four main results. 

First, given the proper form of discounting, either form of KD improves PPL over the CE baseline.

Second, some form of rank-based KD often significantly outperforms KL-based KD.  This is true of 6/9 test-set groups, with only a single experiment showing the reverse (this is a slight conceit, since we are making post-hoc comparisons against a set of PL experiments instead of selecting the best form of discounting during the validation phase). Since, KL-based KD contains both the label order and weight information, it acts as a strong  baseline.  We believe that the reason the rank-based KD often does better is because matching the teacher's ranks is a less strict criterion than matching its probabilities.  It may also be a better criterion when evaluating PPL, since PL may allow for the top ranks to all increase their probability together, so as long as the GT is in the top ranks, then PPL will decrease.  {\bf Importantly, these results also indicate that performance of the $N$-grams will be due to their inherent value as a teacher and will not be hindered by the loss function.}  
The results of the PL and PL-s experiments support the claim that the ordering of the labels provides sufficient semantic information to perform KD and that we can forego label weights (with a caveat addressed later).

Third, the $N$-grams act as a competitive teacher, with only GPT-2 consistently outperforming them. 
It seems plausible that the $N$-grams would surpass the BA models, since both types of KD will restrict how much the BA students can deviate from the baseline CE teacher. With BA and BERT, and LSTM students, the raw PL loss often does worse than the CE baseline.  This is indicative of the PL loss failing when it is easier to optimize the lesser-preference ranks over the GT. Our modified PL loss leads to a modest PPL decrease.  

That the $N$-grams outperform BERT speaks more to BERT failing to act as a good teacher rather than the quality of the $N$-gram models.  For BERT, it was important to use the teacher probabilities (see the KL and PL-t vs PL and PL-s experiments).  We believe that reliance on the teacher's probabilities is indicative of poor ranks where the teacher's sharp probability discounts the lower ranks.  A qualitative analysis of the produced ranks shows that the future information can produce ranks that do not generalize well to a student evaluated on a past-context-only task. For example, BERT can see when a sentence ends and so produces ranks which do not try to continue the sentence where GPT-2 and the $N$-grams will.  Thus, the major caveat mentioned above is that label ordering is sufficient, provided those ranks are informative for the student's task.  The $N$-grams are inherently sparse, since they only derive ranks when supported by the context.  In the future, we would like to consider ways of dynamically selecting the top-$k$ which may fix this discounting issue.  This also continues the desire to model the branching set, which are differently sized for each context.




The fourth result is how well GPT-2 does when paired with the T-XL student, where we see a reduction of almost 10 PPL points.  Compare this with the LSTM student which only sees a reduction of 4 points.  \citet{tang2020understanding} suggested that certain failure cases in KD could be explained by a {\em capacity gap}, where the teacher's distribution is too complex to properly be distilled to the student model.  We believe that the GPT-2 results are the inverse of this phenomenon, where the student falls into a good regime to match the teacher instead of out of the regime.
This makes sense since, outside of the BA models, which may have limited value as teachers, GPT-2 and T-XL are the closest student-teacher pair in terms of specification.   However, since the T-XL model only has $7\%$ more parameters than the LSTM model, we believe this should be considered as a {\em capability gap} where the change in performance between the LSTM and T-XL is more likely due to actual model differences instead of raw capacity.  Additionally, the T-XL model using PL and no discounting was competitive to the discounted forms for BERT and BA and was the best performing model for GPT-2.  This is in contrast to the non-discounted PL using the LSTM student which often did worse than the CE baseline.  This again speaks to a difference in capability between the two types of student models.  We wish to explore this in the future.   


Our results are limited to English datasets\footnote{We explicitly state this limitation to comply with the Bender Rule \citep{DBLP:journals/corr/abs-2110-08300, bender-koller-2020-climbing}.} and there may be generalization issues for aspects of our presented methods for different languages. In particular, we believe that the idea of training a language model via learning to rank is language agnostic; however, the N-gram algorithm for creating the ranks may not be.  For example, an agglutinative language like Turkish, might induce sparsity issues.   Another example is that our N-gram algorithm assumes strong word-order information which will not be true for all languages.  

Rank-based KD is a general idea which allows one to integrate rule-based systems into a student model.  We could improve or generalize the $N$-grams by incorporating syntactic/morphological rules or enforcing co-references when generating the ranks.  This can be done without modifying the model's specification (such as needing to jointly learn the syntax).  Another example of possible uses would to use rank-based KD to integrate an LM into a NMT system by generating target-side ranks, again, without modifying the NMT model.  It may be useful for black-box KD, where one only has access to the teacher's hard predictions  \citep{pmlr-v139-wang21a}, to augment small-data environments with rules,  or to enumerate instructions and answers for instruction fine-tuning \citep{wei2021finetuned}.

In this work, we offer a novel perspective on language modelling by considering it as a multi-label task and develop the methods necessary to train such a model. We connect the idea of LS to ranking via KD by showing that we can incorporate semantic similarity information from an ordered set of words into a LM without requiring associated probabilities.  In our desire to forego using a pre-trained LM, we re-examined how $N$-gram statistics should be incorporated into a neural LM.  We find that this can be done using a simple method which just considers $N$-gram contexts, and that this surpasses the CE baseline and is comparable to using BA and BERT as teachers.  We also find that, for language modelling, a rank-based KD is often a superior method for KD.

\section{Acknowledgements}

We thank our reviewers for their insightful feedback.  

Resources used in preparing this research were provided, in part, by the Province of Ontario, the Government of Canada through CIFAR, and companies sponsoring the Vector Institute. 

We acknowledge the support of the Natural Sciences and Engineering Research Council of Canada (NSERC).

Nous remercions le Conseil de recherches en sciences naturelles et en génie du Canada (CRSNG) de son soutien.

\putbib

\end{bibunit}

\appendix   
\begin{bibunit}
\section{Implementation details and extra experiments}\label{app::imp-details}

\subsection{Teacher models}

We use Hugging Face's implementation of BERT and GPT-2  \citep{wolf-etal-2020-transformers}.  In order to convert the models to a word-level vocabulary, we average the hidden states for each sub-word. For GPT-2, we respected its auto-regressive nature by off-setting the hidden states by one time-step to the past.  We freeze all but the final layer and introduced a word-level embedding table of size $|V| \times h$, where $h$ is the final hidden state size.  These fine-tuned models can be viewed as a sub-word-aware LMs where the input is sub-words and the output is whole words \citep{ling2015finding, kim2016character}.  The ranks are generated by treating the training set as one contiguous text with a conditioning overlap of 50 tokens.  The training set for the BERT models with Wiki02 does not perfectly match the standard training set.  This is because the BERT tokenizer would sometimes fail to map between sub-words and whole-words for words containing a very small set of non-Latin characters.

Instead of including the teacher models directly into training the students, we saved the top-$k$ predictions along with their logits \citep{tan2018multilingual}.  This was necessary, as the Hugging Face models and the student models required different versions of PyTorch (we were unable to replicate the LSTM results in a version of PyTorch $\geq 1.0$).  This also saved computation since we did not need to query the teacher model every batch.

\subsection{Student models}

We use the hyper-parameters from the official implementation for our LSTM models, with the exception that we increased the number of training epochs and when the optimizer switches to exponential averaging.  We did this for the KD experiments in order to account for cycling the KD term.  We do not believe that the CE baseline would have significantly changed with extra epochs.

For our T-XL model, we use 14 layers, 10 heads, a hidden state size of 400, and a feed-forward size of 1000.  We train and evaluate with a target length of 150 tokens and a memory length of 150.  We use Ranger for the optimizer \citep{Ranger}.  We use various forms of variational dropout, which the original authors used when applying T-XL to PTB, but did not use weight averaging \citep{dai-etal-2019-transformer}.
We referenced \href{https://github.com/TimDettmers/transformer-xl/tree/master}{Tim Dettmers's implementation} for the necessary changes required to get T-XL to work with Wiki02.

For the top-$k$ KL-based KD, it is important not to apply temperature smoothing, $\tau$, to the student but to do so for the teacher as in  

\begin{equation} \label{KD}
    L = \alpha \mathrm{CE} + (1-\alpha) \mathrm{KL}(\sigma(w_{\mathrm{kd}}/\tau), \sigma(w_s) ),
\end{equation}

where the overall loss $L$  is linearly  interpolated between the CE and KL (KD) losses using $\alpha$.  
The KL loss tries to match the student logit, $w_s$, against the teacher's logit, $w_{\mathrm{kd}}$.  $\sigma$ is the softmax function.  
 
This is in contrast to traditional KD which applies the same temperature to both the student and the teacher.  However, in our case, since the teacher is smoothed over only the top-$k$ word-types while the student would be smoothed over the full vocabulary, this would make the two distributions very different post-normalization.  We do not multiply the gradients by $\tau^2$.  We also interpolate the PL loss with CE for the PL experiments, though this could also have been achieved by just re-weighing the first term relative to the others to avoid the extra computation cost.   This was done because our initial implementation of PL was numerically unstable, causing NANs.   In these cases, we could back-off to just CE, however, the stability issue was later fixed but the interpolation with CE was not removed.   

Each experiment was hyper-parameter tuned using different ranges of values depending on the given combination of dataset, student, teacher, and loss.  The general ranges of hyper parameters was the value of $k \in \{10, 15, 20\}$, the temperature, $\tau \in [2-16]$ for KL and PL-t experiments and the stepped value $\eta \in [0.2 -0.6]$ for the PL-s experiments.  Cycling was controlled by two hyper-parameters which specified how many epochs constituted a cycle and the minimum value of $\alpha$.  Interpolating with CE re-weighs the original GT relative to the $2, \dots, k$ terms.  However, both $\tau$ and $\eta$ also have this effect, and this also depends on the value of $k$.  This creates a situation where these hyper-parameters are closely coupled and sensitive to each other.  We believe that using annealing instead of cycling may reduce this sensitivity \citep{jafari-etal-2021-annealing}.     

\subsection{Random teacher}

We found that it was important to cycle the $\alpha$ value to trade off between the KD and CE terms.  In order to determine if the decrease in PPL was due to the intrinsic value of the teacher model, and not just an affect of the cycling, we replaced the top-$2$ though $k$ ranks (keeping the original GTs as $k=1$) with random words. Using an LSTM student with the random teacher resulted in a PPL of  63.19 for PTB, and 70.40 for Wiki02 on the test sets.  Since these were both worse than the CE baselines, this indicates that cycling requires a good teacher and is not an independent phenomenon.

\subsection{Pairwise hinge loss}

There are other surrogate rank losses that could be used instead of Plackett-Luce.  We experimented with a pair-wise hinge loss (PWH).  This was selected because PWH can naturally express weak orderings while we needed to modify PL in order to consider them. 
We made use of negative samples in addition to pairs derived from the teacher's top-$k$.  To derive $n$ negative samples, we used the student models own top-$n+k$ predictions cut down to $n$, so as not to include any of the teacher's top-$k$ predictions \citep{pmlr-v130-reddi21a}.  Initial experimentation did not show any benefit to using PWH over PL.  
Using GPT-2 as a teacher with an LSTM student on PTB and the teacher's probabilities for discounting resulted in an average PPL of 55.95 on the test set.  
Combining PWH with PL did see a small reduction in PPL over either individually, with an average PPL of 55.28, however, this increased training time and complicated hyper-parameter tuning since there are now three loss functions to be interpolated.  Using the N-gram teacher with stepped discounting on PTB, achieved an average PPL of 58.27.

\section{Memory efficient algorithm for branching set construction}\label{app::mem}

$N$-grams require a lot of memory since they need to store all possible contexts.   In Algorithm \ref{alg:mem}, we present a memory-efficient algorithm for creating the ranks using multiple passes through the training data.

We can build up the ranks using multiple passes through the dataset, $D$, of $T$ tokens.  Let a context schema be any context of size $p$ past words and $f$ future words (where, in contrast, an order is a specific context of $p$ past words and $f$ future words).  If there are $M$ context schemas, $S$, we use $2 \times M + 1$ passes through the data. In the first phase, we do $M$ passes; one for each schema to collect all the orders of that schema's size.  These are each individually saved to disk so that the memory usage does not grow  significantly larger than the dataset size.    In the second phase, we do a pass to set the ground-truths and then a pass for each of the collected orders to create a $T \times k$ rank matrix, $R$ (with lengths $L$).   This can be efficiently parallelized in two ways. In the first phase, each of the first $M$ passes can be processed separately, so that it does not scale poorly with $M$, given sufficient CPUs.  Then, in the second phase, $R$ can be partitioned and each partition created in parallel.  In practice, there are some tricks necessary to make this efficient, such as using ordered sets for efficient `{\em matrix}' look ups (line 23) and pruning the orders.

\begin{algorithm}
  \caption{Multi-Pass Branching Set Construction
    \label{alg:mem}}
  \begin{algorithmic}[1]
    \Statex
    \Function{process text}{$D, S, k$}
    
    \For{$s \in S $}  \Comment{Schemas are an ordered set}
         \State $C_s \leftarrow \mathrm{dict()}$  \Comment{Context to branching set}
         \For{$t \leftarrow 0, \dots T$}
         \State $c, w_t \leftarrow \mathrm{getContextAndWord}(D[t])$  
         \State $B \leftarrow C_s[c]$  \Comment{Empty set if does not exist}
            \If{$w_t \not\in B$ and $|B| < q -1$ }
    		    \State $B\mathrm{.add}(w_t)$  \Comment{Add $w_t$ to branching set}
    	    \EndIf
         \EndFor
          \State $\mathrm{save}(C_s)$ 
    \EndFor

     \State $R, L \leftarrow \mathrm{zeros}(T, k), \mathrm{ones}(T)$   \Comment{Second phase}
     
         \For{$t \leftarrow 0, \dots T$}
          \State  $R[t, 0] \leftarrow D[t]$   \Comment{Set GT}
          \EndFor

    \For{$C_s \in \{C_1, \dots C_M\} $} 
    \State $\mathrm{load}(C_s)$ 
    \For{$t \leftarrow 0, \dots T$}
        
        \State $c \leftarrow \mathrm{get context}(D[t])$  
				\State $B \leftarrow C_s[c]$
    \For{$w \in B$}  \Comment{For each word in BS}
					\If{$w \not\in R[t, :L[t]]$ }  
					\State  $R[t, L[t]] \leftarrow w$   \Comment{Set word}
					\State $ L[t] \leftarrow L[t] + 1$
                    \EndIf
    \EndFor
    \EndFor
    \EndFor
    \State \Return{$R, L$}
    \EndFunction
  \end{algorithmic}
\end{algorithm}

\section{Vectorized algorithm for Plackett-Luce loss}\label{app::pl-alg}

We present an efficient vectorized algorithm for PL in Algorithm \ref{alg:PLRL}.  This algorithm is batched for a batch size $b$ and sequence length $q$.  $w$ are the logits of shape $[b \times q, |V|]$.  $y$ are the targets of shape $[b \times q, k_m]$ for a max rank of $k_m$.   The targets are padded such that they all have $k_m$ ranks and the padded values will be masked out.  $k$ are the target lengths of shape $[b \times q]$, which will define the mask.  When using the weak ordering modification, we need to know which targets are in the same order. $o$ are the {\em order} values of shape $[b \times q, k_m]$. The order values will be indices in the range $[0, k_m-1]$, which index the first element of an order. $f$ are the values derived from a rank discounting function, i.e. either the teacher's top-$k$ probabilities for PL-t or the stepped values for PL-s.

\begin{algorithm}
  \caption{vectorized Plackett-Luce
    \label{alg:PLRL}}
  \begin{algorithmic}[1]
    \Statex
    \Function{PL}{$w, y, k, o, f$}
    	\State $m \leftarrow \mathrm{max}(w)$ \Comment{For log-sum-exp trick}
    	\State $s \leftarrow e^{w-m}$  \Comment{Scores}
    	\State $w \leftarrow \mathrm{gather}(w, y)$ \Comment{Reorders and cuts to $k_m$}
    	\State $g \leftarrow e^{w-m}$  
    	
    	\Comment{Shift by column of zeros since first term is fully normalized}
    	
    	\State $g \leftarrow \mathrm{cat}(\mathrm{zeros([b \times q, 1])}, g[:, :-1])$ 
    	\State $Z \leftarrow \mathrm{sum}(s, \mathrm{dim}=-1, \mathrm{keepdim})$ 
    	\State $Z_i \leftarrow \mathrm{cumsum}(g, \mathrm{dim}=-1)$
		\If{$o$}  \Comment{ When ranks are weakly ordered}
		
		\Comment{Undoes cumulated sum by duplicating head $Z_i$ value across each order}
		
		    \State $Z_i \leftarrow \mathrm{gather}(Z_i, o)$  
    	\EndIf
    	\State $Z \leftarrow \log(Z - Z_i + 0.00001) + m$ \Comment{ NANs} 
    	\State $l \leftarrow Z - w$  \Comment{Loss}
    	\If{$f$}  \Comment{When using discounting}
    		\State $l \leftarrow l \odot f$  
    	\EndIf
    	\State $l \leftarrow l \odot \mathrm{seqmask}(k)$
      \State \Return{$l$}
      \EndFunction
  \end{algorithmic}
\end{algorithm}

In algorithm \ref{alg:PLRL}, the gather on line 4 accounts for most of the extra computational cost compared to softmax-CE.

\section{Frequency distributions}\label{app::freq-dist}

\begin{figure}[H]
\centerline{  
\title{Log-log plot for Wiki02 using BERT's top-$10$}
\includegraphics[scale=0.35]{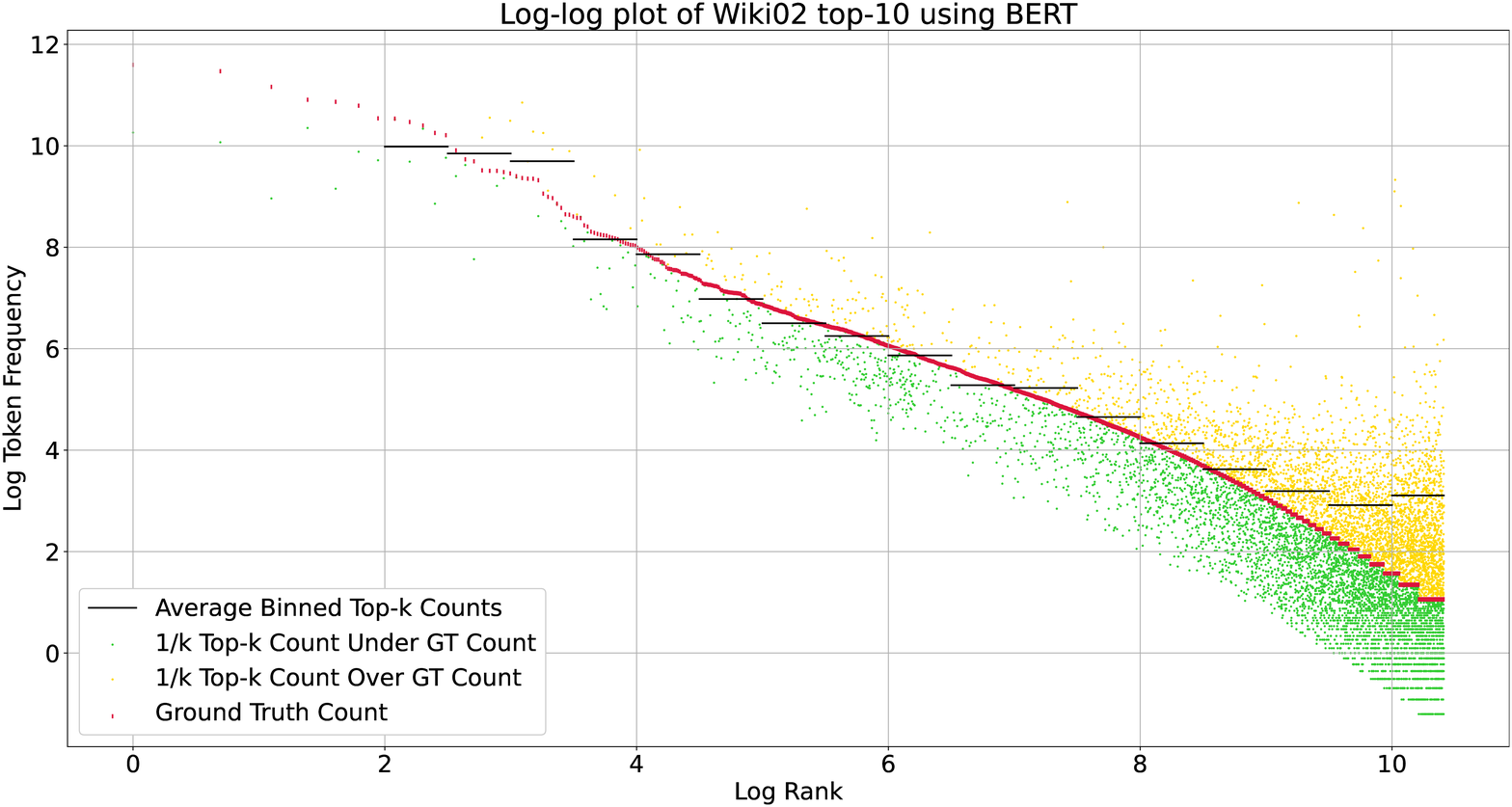}}
\caption{Top-$10$ frequency distribution relative to the original ground-truth distribution for Wiki02 using 24-layer BERT with whole-word-masking.}
\label{fig::stats-wiki02-bert}
\end{figure}

\begin{figure}[H]
\centerline{  
\includegraphics[scale=0.35]{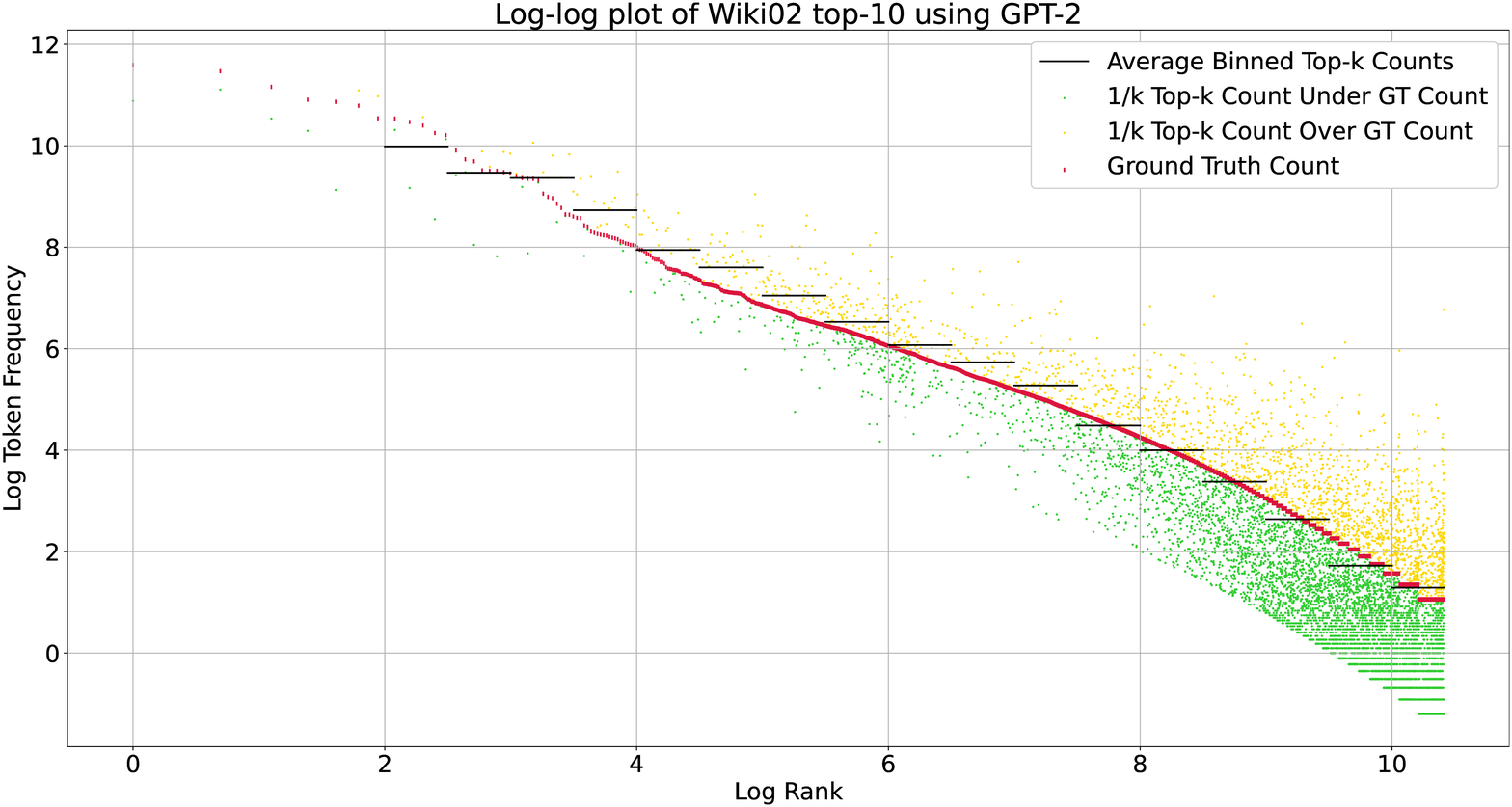}}
\caption{Top-$10$ frequency distribution relative to the original ground-truth distribution for Wiki02 using large GPT-2.}
\label{fig::stats-wiki02-gpt}
\end{figure}

An initial concern was that BERT or GPT-2 might fail to be a good teacher  if the distribution of the top-$k$ ranks was significantly different from ground-truth distribution; this could occur if the high-frequency words became amplified while the low-frequency words diminished.   In order to investigate this,  we plot the log frequency-rank distribution of the original ground-truths (forming a semi-linear downward line) in Figures \ref{fig::stats-wiki02-bert} and  \ref{fig::stats-wiki02-gpt}.  We then plot the $\frac{1}{10}$ of the frequency of the top-$10$ for individual word-types (forming dots above and below the ground-truth line) and binned averages.  This shows that, while individual word-types might be amplified or diminished in relative frequency, the overall binned averages do not stray far from the original distribution.   The figures have been down sampled by a rate of 4.     

\section{Top-$k$ samples}\label{app::topk-samples}

We generate top-$k$ samples using BERT, GPT-2, and the $N$-grams from the Wiki02 training partition at random positions.  The weak orders are grouped for the $N$-grams, where BERT and GPT-2 always produce a strong ordering.  Note that the samples are filtered such that only a limited number of $N$-gram orders are used (so that they can be coloured using a colour-blind friendly scheme) and so that examples only contain Latin characters.      

\begin{table}[H]  \label{tab::Wiki02-sample-465}
\begin{center}
\begin{tabular}{r|rrr|rrr}
Rank & BERT & GPT-2 & N-grams & BERT & GPT-2 & N-grams \\ \hline
1  & \multicolumn{3}{c|}{\emph{debuted}} & \multicolumn{3}{c}{\emph{and}} \\
\hline
2  & debut & peaked & \cellcolor[rgb]{0.843, 0.188, 0.153}{peaked} & to & at & \cellcolor[rgb]{0.843, 0.188, 0.153}{at} \\
3  & debuting & reached & \cellcolor[rgb]{0.843, 0.188, 0.153}{was} & which & atop & \cellcolor[rgb]{0.988, 0.553, 0.349}{on} \\
4  & charted & is &  & as & with & \cellcolor[rgb]{0.988, 0.553, 0.349}{in} \\
5  & peaked & appeared &  & for & as & \cellcolor[rgb]{0.988, 0.553, 0.349}{to} \\
6  & premiered & entered &  & at & on &  \\
7  & topped & ranked &  & And & in &  \\
8  & cursed & became &  & by & from &  \\
9  & rookie & was &  & from & to &  \\
10  & Deacon & has &  & who & , &  \\
\hline
1  & \multicolumn{3}{c|}{\emph{peaked}} & \multicolumn{3}{c}{\emph{at}} \\
\hline
2  & charted & stayed & \cellcolor[rgb]{0.843, 0.188, 0.153}{finally} & At & atop & \cellcolor[rgb]{0.843, 0.188, 0.153}{on} \\
3  & peaking & remained &  & as & in & \cellcolor[rgb]{0.988, 0.553, 0.349}{atop} \\
4  & reached & reached &  & to & within & \cellcolor[rgb]{0.988, 0.553, 0.349}{within} \\
5  & Peaking & spent &  & from & with &  \\
6  & peak & entered &  & for & as &  \\
7  & singled & debuted &  & on & on &  \\
8  & filed & appeared &  & by & from &  \\
9  & grossed & moved &  & reached & between &  \\
10  & topped & became &  & with & number &  \\
\hline
1  & \multicolumn{3}{c|}{\emph{number}} & \multicolumn{3}{c}{\emph{89}} \\
\hline
2  & Number & numbers & \cellcolor[rgb]{0.843, 0.188, 0.153}{No.} & 9 & 39 & \cellcolor[rgb]{0.843, 0.188, 0.153}{nine} \\
3  & top & \# &  & 1989 & 33 & \cellcolor[rgb]{0.988, 0.553, 0.349}{15} \\
4  & figure & the &  & nine & 26 & \cellcolor[rgb]{0.784, 0.745, 0.565}{32} \\
5  & least & its &  & 1789 & 12 & \cellcolor[rgb]{0.784, 0.745, 0.565}{11} \\
6  & nine & no &  & 9th & 23 & \cellcolor[rgb]{0.784, 0.745, 0.565}{31} \\
7  & Top & position &  & 09 & 5 & \cellcolor[rgb]{0.784, 0.824, 0.824}{25} \\
8  & Figure & 39 &  & 1889 & 20 & \cellcolor[rgb]{0.569, 0.749, 0.859}{eight} \\
9  & member & chart &  & 179 & twelve & \cellcolor[rgb]{0.569, 0.749, 0.859}{18} \\
10  & chart & a &  & 90 & 76 & \cellcolor[rgb]{0.271, 0.459, 0.706}{84} \\
\hline
1  & \multicolumn{3}{c|}{\emph{in}} & \multicolumn{3}{c}{\emph{the}} \\
\hline
2  & In & on & \cellcolor[rgb]{0.843, 0.188, 0.153}{on} & The & its & \cellcolor[rgb]{0.843, 0.188, 0.153}{)} \\
3  & Iny & , &  & THE & July & \cellcolor[rgb]{0.988, 0.553, 0.349}{every} \\
4  & IN & . &  & ( & October & \cellcolor[rgb]{0.988, 0.553, 0.349}{this} \\
5  & on & and &  & which & September & \cellcolor[rgb]{0.988, 0.553, 0.349}{an} \\
6  & by & with &  & an & June &  \\
7  & for & ( &  & their & April &  \\
8  & from & ; &  & its & November &  \\
9  & On & for &  & was & August &  \\
10  & his & at &  & of & December &  \\
\hline
\end{tabular}
\end{center}
\caption{Example top-k from bert-wwm-24, gpt2-774, and N-grams using the Wiki02 training partition at block 465 with ground-truth text `debuted and peaked at number 89 in the'.  }
\end{table}


\begin{table}[H]  \label{tab::Wiki02-sample-558}
\begin{center}
\begin{tabular}{r|rrr|rrr}
Rank & BERT & GPT2 & Ngrams & BERT & GPT2 & Ngrams \\ \hline
1  & \multicolumn{3}{c|}{\emph{organisers}} & \multicolumn{3}{c}{\emph{to}} \\
\hline
2  & organiser & government & \cellcolor[rgb]{0.843, 0.188, 0.153}{ships} & To & and & \cellcolor[rgb]{0.843, 0.188, 0.153}{have} \\
3  & mourners & authorities &  & who & for & \cellcolor[rgb]{0.843, 0.188, 0.153}{.} \\
4  & rescuers & company &  & by & of & \cellcolor[rgb]{0.843, 0.188, 0.153}{adapted} \\
5  & Templers & police &  & from & , &  \\
6  & chasers & Government &  & , & with &  \\
7  & anglers & and &  & could & about &  \\
8  & chroniclers & propaganda &  & an & in &  \\
9  & taker & Parliament &  & as & who &  \\
10  & admirers & media &  & would & . &  \\
\hline
1  & \multicolumn{3}{c|}{\emph{suggest}} & \multicolumn{3}{c}{\emph{ending}} \\
\hline
2  & suggestion & raise & \cellcolor[rgb]{0.843, 0.188, 0.153}{the} & Ending & that & \cellcolor[rgb]{0.843, 0.188, 0.153}{that} \\
3  & ask & produce &  & end & a & \cellcolor[rgb]{0.988, 0.553, 0.349}{to} \\
4  & consider & remake &  & ended & the & \cellcolor[rgb]{0.988, 0.553, 0.349}{meanings} \\
5  & question & be &  & ends & an & \cellcolor[rgb]{0.988, 0.553, 0.349}{a} \\
6  & recommend & have &  & leaving & they & \cellcolor[rgb]{0.988, 0.553, 0.349}{she} \\
7  & propose & extend &  & holding & to & \cellcolor[rgb]{0.988, 0.553, 0.349}{amendments} \\
8  & suggested & play &  & closing & further & \cellcolor[rgb]{0.988, 0.553, 0.349}{original} \\
9  & recommendation & replace &  & eliminating & it & \cellcolor[rgb]{0.988, 0.553, 0.349}{Meyerbeer} \\
10  & offer & provide &  & finishing & extending &  \\
\hline
1  & \multicolumn{3}{c|}{\emph{the}} & \multicolumn{3}{c}{\emph{ceremony}} \\
\hline
2  & The & their &  & event & event & \cellcolor[rgb]{0.843, 0.188, 0.153}{song} \\
3  & to & this &  & ceremonies & war & \cellcolor[rgb]{0.843, 0.188, 0.153}{knuckleball} \\
4  & of & it &  & festival & service & \cellcolor[rgb]{0.843, 0.188, 0.153}{criminal} \\
5  & THE & all &  & ceremonial & practice & \cellcolor[rgb]{0.843, 0.188, 0.153}{track} \\
6  & from & these &  & process & annual & \cellcolor[rgb]{0.843, 0.188, 0.153}{dead} \\
7  & an & its &  & site & anniversary & \cellcolor[rgb]{0.843, 0.188, 0.153}{island} \\
8  & by & a &  & speech & tradition &  \\
9  & for & an &  & display & battle &  \\
10  & their & of &  & held & operation &  \\
\hline
1  & \multicolumn{3}{c|}{\emph{,}} & \multicolumn{3}{c}{\emph{believing}} \\
\hline
2  & ) & in & \cellcolor[rgb]{0.843, 0.188, 0.153}{.} & Believing & and & \cellcolor[rgb]{0.843, 0.188, 0.153}{the} \\
3  & to & . & \cellcolor[rgb]{0.843, 0.188, 0.153}{officially} & believed & so & \cellcolor[rgb]{0.843, 0.188, 0.153}{one} \\
4  & who & at & \cellcolor[rgb]{0.843, 0.188, 0.153}{;} & fearing & but & \cellcolor[rgb]{0.843, 0.188, 0.153}{Obama} \\
5  & as & and & \cellcolor[rgb]{0.843, 0.188, 0.153}{that} & expecting & as & \cellcolor[rgb]{0.843, 0.188, 0.153}{Roosevelt} \\
6  & for & on & \cellcolor[rgb]{0.843, 0.188, 0.153}{starting} & believe & because & \cellcolor[rgb]{0.988, 0.553, 0.349}{and} \\
7  & by & to & \cellcolor[rgb]{0.843, 0.188, 0.153}{continues} & believes & stating & \cellcolor[rgb]{0.988, 0.553, 0.349}{commemorating} \\
8  & at & with & \cellcolor[rgb]{0.843, 0.188, 0.153}{time} & belief & which & \cellcolor[rgb]{0.784, 0.745, 0.565}{survived} \\
9  & and & there &  & Holding & to &  \\
10  & that & so &  & seeing & however &  \\
\hline
\end{tabular}
\end{center}
\caption{Example top-k from bert-wwm-24, gpt2-774, and N-grams using the Wiki02 training partition at block 558 with ground-truth text `organisers to suggest ending the ceremony , believing'.  }
\end{table}

\begin{table}[H]  \label{tab::Wiki02-sample-559}
\begin{center}
\begin{tabular}{r|rrr|rrr}
Rank & BERT & GPT2 & Ngrams & BERT & GPT2 & Ngrams \\ \hline
1  & \multicolumn{3}{c|}{\emph{became}} & \multicolumn{3}{c}{\emph{an}} \\
\hline
2  & become & is & \cellcolor[rgb]{0.843, 0.188, 0.153}{was} & An & a &  \\
3  & gained & , & \cellcolor[rgb]{0.843, 0.188, 0.153}{up} & a & apparent &  \\
4  & followed & was &  & A & more &  \\
5  & becomes & and &  & one & very &  \\
6  & saw & in &  & as & the &  \\
7  & remained & of &  & that & so &  \\
8  & becoming & has &  & something & necessary &  \\
9  & made & to &  & another & significant &  \\
10  & ended & from &  & to & one &  \\
\hline
1  & \multicolumn{3}{c|}{\emph{issue}} & \multicolumn{3}{c}{\emph{in}} \\
\hline
2  & idea & important & \cellcolor[rgb]{0.843, 0.188, 0.153}{agnostic} & In & when & \cellcolor[rgb]{0.843, 0.188, 0.153}{that} \\
3  & problem & integral & \cellcolor[rgb]{0.988, 0.553, 0.349}{Academy} & for & for & \cellcolor[rgb]{0.988, 0.553, 0.349}{,} \\
4  & issues & entity &  & with & , & \cellcolor[rgb]{0.988, 0.553, 0.349}{of} \\
5  & area & apparent &  & from & because & \cellcolor[rgb]{0.988, 0.553, 0.349}{still} \\
6  & amount & obvious &  & on & of & \cellcolor[rgb]{0.988, 0.553, 0.349}{about} \\
7  & important & idea &  & by & . & \cellcolor[rgb]{0.988, 0.553, 0.349}{for} \\
8  & obstacle & essential &  & IN & with & \cellcolor[rgb]{0.988, 0.553, 0.349}{with} \\
9  & event & indicator &  & that & due & \cellcolor[rgb]{0.988, 0.553, 0.349}{due} \\
10  & example & influence &  & and & to &  \\
\hline
1  & \multicolumn{3}{c|}{\emph{later}} & \multicolumn{3}{c}{\emph{games}} \\
\hline
2  & Later & the & \cellcolor[rgb]{0.843, 0.188, 0.153}{the} & game & episodes & \cellcolor[rgb]{0.843, 0.188, 0.153}{decades} \\
3  & late & a & \cellcolor[rgb]{0.843, 0.188, 0.153}{May} & Games & releases & \cellcolor[rgb]{0.843, 0.188, 0.153}{performances} \\
4  & rest & some & \cellcolor[rgb]{0.843, 0.188, 0.153}{Jifna} & players & editions & \cellcolor[rgb]{0.843, 0.188, 0.153}{years} \\
5  & future & an & \cellcolor[rgb]{0.843, 0.188, 0.153}{2008} & Game & stages & \cellcolor[rgb]{0.843, 0.188, 0.153}{series} \\
6  & Late & its & \cellcolor[rgb]{0.843, 0.188, 0.153}{Haiti} & played & works & \cellcolor[rgb]{0.843, 0.188, 0.153}{life} \\
7  & many & Japanese & \cellcolor[rgb]{0.843, 0.188, 0.153}{an} & areas & versions & \cellcolor[rgb]{0.843, 0.188, 0.153}{times} \\
8  & other & terms &  & laws & series &  \\
9  & second & that &  & ships & years &  \\
10  & said & Japan &  & shows & parts &  \\
\hline
1  & \multicolumn{3}{c|}{\emph{,}} & \multicolumn{3}{c}{\emph{where}} \\
\hline
2  & for & . &  & Where & and & \cellcolor[rgb]{0.843, 0.188, 0.153}{a} \\
3  & ) & in &  & there & so & \cellcolor[rgb]{0.988, 0.553, 0.349}{marketed} \\
4  & that & and &  & and & as & \cellcolor[rgb]{0.988, 0.553, 0.349}{causing} \\
5  & ( & as &  & who & because & \cellcolor[rgb]{0.988, 0.553, 0.349}{in} \\
6  & who & when &  & There & when & \cellcolor[rgb]{0.988, 0.553, 0.349}{but} \\
7  & : & because &  & with & however & \cellcolor[rgb]{0.988, 0.553, 0.349}{still} \\
8  & " & : &  & as & due & \cellcolor[rgb]{0.988, 0.553, 0.349}{was} \\
9  & as & ' &  & both & particularly & \cellcolor[rgb]{0.988, 0.553, 0.349}{including} \\
10  & from & of &  & showing & but & \cellcolor[rgb]{0.988, 0.553, 0.349}{with} \\
\hline
\end{tabular}
\end{center}
\caption{Example top-k from bert-wwm-24, gpt2-774, and N-grams using the Wiki02 training partition at block 559 with ground-truth text `became an issue in later games , where'.  }
\end{table}

\begin{table}[H]  \label{tab::Wiki02-sample-436}
\begin{center}
\begin{tabular}{r|rrr|rrr}
Rank & BERT & GPT2 & Ngrams & BERT & GPT2 & Ngrams \\ \hline
1  & \multicolumn{3}{c|}{\emph{is}} & \multicolumn{3}{c}{\emph{shown}} \\
\hline
2  & was & can &  & showed & able & \cellcolor[rgb]{0.843, 0.188, 0.153}{unemployed} \\
3  & does & has &  & seen & capable & \cellcolor[rgb]{0.988, 0.553, 0.349}{famed} \\
4  & has & does &  & showing & aware & \cellcolor[rgb]{0.988, 0.553, 0.349}{<unk>} \\
5  & Is & appears &  & played & a & \cellcolor[rgb]{0.988, 0.553, 0.349}{late} \\
6  & as & may &  & given & also & \cellcolor[rgb]{0.988, 0.553, 0.349}{helpless} \\
7  & are & makes &  & used & said & \cellcolor[rgb]{0.988, 0.553, 0.349}{summoned} \\
8  & goes & uses &  & found & physically & \cellcolor[rgb]{0.988, 0.553, 0.349}{able} \\
9  & also & also &  & thought & not & \cellcolor[rgb]{0.988, 0.553, 0.349}{unable} \\
10  & uses & will &  & tried & still &  \\
\hline
1  & \multicolumn{3}{c|}{\emph{to}} & \multicolumn{3}{c}{\emph{be}} \\
\hline
2  & To & in & \cellcolor[rgb]{0.843, 0.188, 0.153}{surrounded} & Be & have & \cellcolor[rgb]{0.843, 0.188, 0.153}{the} \\
3  & as & as &  & are & speak & \cellcolor[rgb]{0.843, 0.188, 0.153}{have} \\
4  & that & capable &  & remain & possess &  \\
5  & a & being &  & keep & enjoy &  \\
6  & can & using &  & ask & do &  \\
7  & by & with &  & know & experience &  \\
8  & who & attempting &  & have & know &  \\
9  & could & having &  & continue & also &  \\
10  & an & producing &  & believe & take &  \\
\hline
1  & \multicolumn{3}{c|}{\emph{curious}} & \multicolumn{3}{c}{\emph{about}} \\
\hline
2  & wondering & capable & \cellcolor[rgb]{0.843, 0.188, 0.153}{<unk>} & About & and & \cellcolor[rgb]{0.843, 0.188, 0.153}{collection} \\
3  & questions & a & \cellcolor[rgb]{0.843, 0.188, 0.153}{precise} & asking & , & \cellcolor[rgb]{0.843, 0.188, 0.153}{decision} \\
4  & wondered & able & \cellcolor[rgb]{0.988, 0.553, 0.349}{mad} & into & of & \cellcolor[rgb]{0.843, 0.188, 0.153}{.} \\
5  & curiosity & aware & \cellcolor[rgb]{0.988, 0.553, 0.349}{serious} & using & in & \cellcolor[rgb]{0.843, 0.188, 0.153}{fact} \\
6  & Curious & physically & \cellcolor[rgb]{0.988, 0.553, 0.349}{born} & questions & when & \cellcolor[rgb]{0.843, 0.188, 0.153}{to} \\
7  & amazed & an & \cellcolor[rgb]{0.988, 0.553, 0.349}{less} & for & . & \cellcolor[rgb]{0.843, 0.188, 0.153}{features} \\
8  & suspicious & very & \cellcolor[rgb]{0.988, 0.553, 0.349}{honest} & under & to & \cellcolor[rgb]{0.843, 0.188, 0.153}{arts} \\
9  & interesting & intelligent &  & " & as & \cellcolor[rgb]{0.843, 0.188, 0.153}{by} \\
10  & jealous & sensitive &  & over & at &  \\
\hline
1  & \multicolumn{3}{c|}{\emph{the}} & \multicolumn{3}{c}{\emph{world}} \\
\hline
2  & The & them &  & life & evil &  \\
3  & 's & her &  & World & emotions &  \\
4  & ) & a &  & worlds & supernatural &  \\
5  & a & human &  & Life & life &  \\
6  & ( & and &  & job & human &  \\
7  & " & their &  & Book & nature &  \\
8  & THE & humans &  & way & truth &  \\
9  & , & what &  & man & paranormal &  \\
10  & their & how &  & lives & lives &  \\
\hline
\end{tabular}
\end{center}
\caption{Example top-k from bert-wwm-24, gpt2-774, and N-grams using the Wiki02 training partition at block 436 with ground-truth text `is shown to be curious about the world'.  }
\end{table}

\begin{table}[H]  \label{tab::Wiki02-sample-130}
\begin{center}
\begin{tabular}{r|rrr|rrr}
Rank & BERT & GPT2 & Ngrams & BERT & GPT2 & Ngrams \\ \hline
1  & \multicolumn{3}{c|}{\emph{and}} & \multicolumn{3}{c}{\emph{the}} \\
\hline
2  & And & leaving & \cellcolor[rgb]{0.843, 0.188, 0.153}{the} & The & it & \cellcolor[rgb]{0.843, 0.188, 0.153}{about} \\
3  & who & so & \cellcolor[rgb]{0.988, 0.553, 0.349}{was} & been & a & \cellcolor[rgb]{0.988, 0.553, 0.349}{their} \\
4  & but & taking & \cellcolor[rgb]{0.988, 0.553, 0.349}{reaching} & its & by & \cellcolor[rgb]{0.988, 0.553, 0.349}{was} \\
5  & as & leading &  & a & only & \cellcolor[rgb]{0.988, 0.553, 0.349}{its} \\
6  & which & with &  & was & then & \cellcolor[rgb]{0.988, 0.553, 0.349}{her} \\
7  & for & which &  & their & when & \cellcolor[rgb]{0.988, 0.553, 0.349}{his} \\
8  & with & allowing &  & were & there & \cellcolor[rgb]{0.988, 0.553, 0.349}{at} \\
9  & by & when &  & an & in &  \\
10  & where & thereby &  & for & were &  \\
\hline
1  & \multicolumn{3}{c|}{\emph{first}} & \multicolumn{3}{c}{\emph{service}} \\
\hline
2  & First & restaurant & \cellcolor[rgb]{0.843, 0.188, 0.153}{civil} & Service & restaurant & \cellcolor[rgb]{0.843, 0.188, 0.153}{flight} \\
3  & main & new & \cellcolor[rgb]{0.988, 0.553, 0.349}{silver} & load & food & \cellcolor[rgb]{0.988, 0.553, 0.349}{ship} \\
4  & major & dining & \cellcolor[rgb]{0.784, 0.745, 0.565}{military} & services & dining & \cellcolor[rgb]{0.988, 0.553, 0.349}{advisory} \\
5  & only & food & \cellcolor[rgb]{0.784, 0.745, 0.565}{<unk>} & serve & of &  \\
6  & new & opening & \cellcolor[rgb]{0.784, 0.745, 0.565}{main} & news & visitors &  \\
7  & many & kitchen & \cellcolor[rgb]{0.784, 0.745, 0.565}{protection} & military & chef &  \\
8  & final & final &  & served & day &  \\
9  & one & original &  & pair & night &  \\
10  & beginning & work &  & head & pair &  \\
\hline
1  & \multicolumn{3}{c|}{\emph{was}} & \multicolumn{3}{c}{\emph{run}} \\
\hline
2  & were & of &  & ran & only & \cellcolor[rgb]{0.843, 0.188, 0.153}{held} \\
3  & been & on &  & Run & carried & \cellcolor[rgb]{0.843, 0.188, 0.153}{estimated} \\
4  & had & resumed &  & shown & held &  \\
5  & be & had &  & worn & stopped &  \\
6  & could & to &  & used & not &  \\
7  & being & in &  & seen & just &  \\
8  & saw & started &  & read & a &  \\
9  & would & did &  & runs & initiated &  \\
10  & only & at &  & broadcast & completed &  \\
\hline
1  & \multicolumn{3}{c|}{\emph{at}} & \multicolumn{3}{c}{\emph{7}} \\
\hline
2  & At & in & \cellcolor[rgb]{0.843, 0.188, 0.153}{.} & seven & 12 & \cellcolor[rgb]{0.843, 0.188, 0.153}{the} \\
3  & from & just & \cellcolor[rgb]{0.843, 0.188, 0.153}{under} & 7th & 1 & \cellcolor[rgb]{0.843, 0.188, 0.153}{reduced} \\
4  & for & by & \cellcolor[rgb]{0.843, 0.188, 0.153}{over} & seventh & 5 & \cellcolor[rgb]{0.843, 0.188, 0.153}{a} \\
5  & by & on & \cellcolor[rgb]{0.843, 0.188, 0.153}{with} & 07 & 10 & \cellcolor[rgb]{0.843, 0.188, 0.153}{37} \\
6  & on & from & \cellcolor[rgb]{0.843, 0.188, 0.153}{by} & Seven & 4 & \cellcolor[rgb]{0.843, 0.188, 0.153}{up} \\
7  & with & only &  & 700 & 2 & \cellcolor[rgb]{0.988, 0.553, 0.349}{9} \\
8  & as & a &  & 8 & 14 & \cellcolor[rgb]{0.988, 0.553, 0.349}{11} \\
9  & to & for &  & 17 & around &  \\
10  & in & almost &  & 27 & the &  \\
\hline
\end{tabular}
\end{center}
\caption{Example top-k from bert-wwm-24, gpt2-774, and N-grams using the Wiki02 training partition at block 130 with ground-truth text `and the first service was run at 7'.  }
\end{table}

\begin{table}[H]  \label{tab::Wiki02-sample-554}
\begin{center}
\begin{tabular}{r|rrr|rrr}
Rank & BERT & GPT2 & Ngrams & BERT & GPT2 & Ngrams \\ \hline
1  & \multicolumn{3}{c|}{\emph{Museum}} & \multicolumn{3}{c}{\emph{.}} \\
\hline
2  & museum & Channel & \cellcolor[rgb]{0.843, 0.188, 0.153}{Commonwealth} & <eos> & , &  \\
3  & Louvre & Library & \cellcolor[rgb]{0.843, 0.188, 0.153}{citizens} & I. & in &  \\
4  & Smithsonian & Short & \cellcolor[rgb]{0.843, 0.188, 0.153}{battlecruisers} & X. & and &  \\
5  & Gallery & International & \cellcolor[rgb]{0.843, 0.188, 0.153}{protectorate} & Z. & on &  \\
6  & Zoo & Film &  & U. & ( &  \\
7  & Park & Independent &  & E. & of &  \\
8  & museums & Academy &  & Q. & at &  \\
9  & park & Kids &  & C. & for &  \\
10  & Empire & Television &  & = & with &  \\
\hline
1  & \multicolumn{3}{c|}{\emph{The}} & \multicolumn{3}{c}{\emph{DVD}} \\
\hline
2  & the & It & \cellcolor[rgb]{0.843, 0.188, 0.153}{On} & Disc & film & \cellcolor[rgb]{0.843, 0.188, 0.153}{rendering} \\
3  & a & In & \cellcolor[rgb]{0.843, 0.188, 0.153}{<unk>} & DVDs & anime & \cellcolor[rgb]{0.988, 0.553, 0.349}{column} \\
4  & THE & A & \cellcolor[rgb]{0.843, 0.188, 0.153}{Mathews} & overseas & episode & \cellcolor[rgb]{0.988, 0.553, 0.349}{Department} \\
5  & its & As & \cellcolor[rgb]{0.843, 0.188, 0.153}{<eos>} & CD & series & \cellcolor[rgb]{0.988, 0.553, 0.349}{railway} \\
6  & A & " & \cellcolor[rgb]{0.843, 0.188, 0.153}{His} & box & BBC & \cellcolor[rgb]{0.988, 0.553, 0.349}{Strand} \\
7  & an & Other & \cellcolor[rgb]{0.843, 0.188, 0.153}{He} & video & first & \cellcolor[rgb]{0.988, 0.553, 0.349}{National} \\
8  & ( & This & \cellcolor[rgb]{0.988, 0.553, 0.349}{TV} & Steiner & original & \cellcolor[rgb]{0.784, 0.745, 0.565}{Dodgers} \\
9  & Its & For & \cellcolor[rgb]{0.988, 0.553, 0.349}{"} & Hal & English & \cellcolor[rgb]{0.784, 0.745, 0.565}{WWF} \\
10  & of & Some & \cellcolor[rgb]{0.988, 0.553, 0.349}{A} & PAS & movie & \cellcolor[rgb]{0.784, 0.745, 0.565}{band} \\
\hline
1  & \multicolumn{3}{c|}{\emph{released}} & \multicolumn{3}{c}{\emph{ranked}} \\
\hline
2  & release & of & \cellcolor[rgb]{0.843, 0.188, 0.153}{received} & charted & in &  \\
3  & Released & and & \cellcolor[rgb]{0.843, 0.188, 0.153}{contained} & featured & by &  \\
4  & releases & version & \cellcolor[rgb]{0.843, 0.188, 0.153}{included} & rose & on &  \\
5  & releasing & release & \cellcolor[rgb]{0.843, 0.188, 0.153}{called} & appeared & for &  \\
6  & stated & edition & \cellcolor[rgb]{0.843, 0.188, 0.153}{was} & reached & with &  \\
7  & version & was & \cellcolor[rgb]{0.843, 0.188, 0.153}{has} & remained & to &  \\
8  & published & is & \cellcolor[rgb]{0.843, 0.188, 0.153}{version} & finished & as &  \\
9  & election & has &  & Ranked & a &  \\
10  & elected & includes &  & marked & the &  \\
\hline
1  & \multicolumn{3}{c|}{\emph{No.}} & \multicolumn{3}{c}{\emph{1}} \\
\hline
2  & top & \# &  & 1st & 5 & \cellcolor[rgb]{0.843, 0.188, 0.153}{26} \\
3  & no & 39 &  & one & 39 & \cellcolor[rgb]{0.988, 0.553, 0.349}{2} \\
4  & than & 10 &  & One & 7 & \cellcolor[rgb]{0.784, 0.745, 0.565}{5} \\
5  & up & 23 &  & first & 12 & \cellcolor[rgb]{0.784, 0.745, 0.565}{4} \\
6  & to & seventh &  & top & 8 & \cellcolor[rgb]{0.784, 0.745, 0.565}{6} \\
7  & at & 76 &  & up & 26 & \cellcolor[rgb]{0.784, 0.745, 0.565}{8} \\
8  & best & 5 &  & single & 4 & \cellcolor[rgb]{0.784, 0.745, 0.565}{15} \\
9  & for & sixth &  & well & 10 &  \\
10  & as & fourth &  & leader & 76 &  \\
\hline
\end{tabular}
\end{center}
\caption{Example top-k from bert-wwm-24, gpt2-774, and N-grams using the Wiki02 training partition at block 554 with ground-truth text `Museum . The DVD released ranked No. 1'.  }
\end{table}

\begin{table}[H]  \label{tab::Wiki02-sample-75}
\begin{center}
\begin{tabular}{r|rrr|rrr}
Rank & BERT & GPT2 & Ngrams & BERT & GPT2 & Ngrams \\ \hline
1  & \multicolumn{3}{c|}{\emph{Bear}} & \multicolumn{3}{c}{\emph{River}} \\
\hline
2  & Cache & highway & \cellcolor[rgb]{0.843, 0.188, 0.153}{Cass} & river & Creek & \cellcolor[rgb]{0.843, 0.188, 0.153}{<unk>} \\
3  & bear & state & \cellcolor[rgb]{0.988, 0.553, 0.349}{<unk>} & Rivers & Mountain & \cellcolor[rgb]{0.988, 0.553, 0.349}{Creek} \\
4  & Cub & city & \cellcolor[rgb]{0.784, 0.745, 0.565}{Huron} & Pit & Eagle & \cellcolor[rgb]{0.988, 0.553, 0.349}{Lake} \\
5  & Salmon & town & \cellcolor[rgb]{0.784, 0.745, 0.565}{River} & revue & Village & \cellcolor[rgb]{0.988, 0.553, 0.349}{Creeks} \\
6  & book & northern & \cellcolor[rgb]{0.784, 0.745, 0.565}{road} & Creek & Road & \cellcolor[rgb]{0.988, 0.553, 0.349}{Valley} \\
7  & Park & US & \cellcolor[rgb]{0.784, 0.745, 0.565}{Trent} & Launcher & Canyon & \cellcolor[rgb]{0.988, 0.553, 0.349}{)} \\
8  & Snake & county & \cellcolor[rgb]{0.784, 0.745, 0.565}{Mulberry} & rivers & Run & \cellcolor[rgb]{0.988, 0.553, 0.349}{Bryant} \\
9  & Russia & southern &  & Valley & Island &  \\
10  & Bears & East &  & water & National &  \\
\hline
1  & \multicolumn{3}{c|}{\emph{and}} & \multicolumn{3}{c}{\emph{continues}} \\
\hline
2  & as & , & \cellcolor[rgb]{0.843, 0.188, 0.153}{was} & Continuing & follows & \cellcolor[rgb]{0.843, 0.188, 0.153}{Ireland} \\
3  & who & from & \cellcolor[rgb]{0.843, 0.188, 0.153}{in} & continuing & the & \cellcolor[rgb]{0.988, 0.553, 0.349}{the} \\
4  & from & at & \cellcolor[rgb]{0.988, 0.553, 0.349}{Avenue} & continued & proceeds & \cellcolor[rgb]{0.988, 0.553, 0.349}{stretches} \\
5  & And & just &  & continue & passes & \cellcolor[rgb]{0.988, 0.553, 0.349}{then} \\
6  & for & in &  & remains & then & \cellcolor[rgb]{0.988, 0.553, 0.349}{Oghratina} \\
7  & to & to &  & further & is & \cellcolor[rgb]{0.988, 0.553, 0.349}{later} \\
8  & at & on &  & still & parallels & \cellcolor[rgb]{0.988, 0.553, 0.349}{further} \\
9  & by & between &  & Farther & becomes &  \\
10  & in & into &  & keeps & runs &  \\
\hline
1  & \multicolumn{3}{c|}{\emph{east}} & \multicolumn{3}{c}{\emph{through}} \\
\hline
2  & East & through & \cellcolor[rgb]{0.843, 0.188, 0.153}{southeasterly} & Through & along & \cellcolor[rgb]{0.843, 0.188, 0.153}{onto} \\
3  & eastern & north & \cellcolor[rgb]{0.843, 0.188, 0.153}{on} & into & eastwards & \cellcolor[rgb]{0.843, 0.188, 0.153}{to} \\
4  & Eastern & to &  & for & on & \cellcolor[rgb]{0.988, 0.553, 0.349}{past} \\
5  & eastward & along &  & across & toward & \cellcolor[rgb]{0.988, 0.553, 0.349}{of} \\
6  & E & on &  & throughout & to &  \\
7  & article & south &  & in & , &  \\
8  & arch & over &  & down & into &  \\
9  & all & west &  & from & westward &  \\
10  & red & toward &  & though & for &  \\
\hline
1  & \multicolumn{3}{c|}{\emph{rural}} & \multicolumn{3}{c}{\emph{Cache}} \\
\hline
2  & Rural & the & \cellcolor[rgb]{0.843, 0.188, 0.153}{Europe} & Benton & areas & \cellcolor[rgb]{0.843, 0.188, 0.153}{farmland} \\
3  & white & downtown & \cellcolor[rgb]{0.843, 0.188, 0.153}{the} & Bear & and & \cellcolor[rgb]{0.843, 0.188, 0.153}{farm} \\
4  & countryside & a & \cellcolor[rgb]{0.843, 0.188, 0.153}{Des} & Emmett & farm &  \\
5  & quiet & residential & \cellcolor[rgb]{0.843, 0.188, 0.153}{a} & Boulder & Kent &  \\
6  & touring & an &  & Salmon & southern &  \\
7  & central & southern &  & Cascade & residential &  \\
8  & retreating & suburban &  & Cornish & , &  \\
9  & Midwest & farmland &  & Salford & forests &  \\
10  & outdoor & town &  & Milk & County &  \\
\hline
\end{tabular}
\end{center}
\caption{Example top-k from bert-wwm-24, gpt2-774, and N-grams using the Wiki02 training partition at block 75 with ground-truth text `Bear River and continues east through rural Cache'.  }
\end{table}

\begin{table}[H]  \label{tab::Wiki02-sample-102}
\begin{center}
\begin{tabular}{r|rrr|rrr}
Rank & BERT & GPT2 & Ngrams & BERT & GPT2 & Ngrams \\ \hline
1  & \multicolumn{3}{c|}{\emph{at}} & \multicolumn{3}{c}{\emph{the}} \\
\hline
2  & At & into & \cellcolor[rgb]{0.843, 0.188, 0.153}{and} & The & it &  \\
3  & against & from & \cellcolor[rgb]{0.988, 0.553, 0.349}{,} & THE & him &  \\
4  & towards & and & \cellcolor[rgb]{0.988, 0.553, 0.349}{.} & by & his &  \\
5  & as & in & \cellcolor[rgb]{0.988, 0.553, 0.349}{also} & with & Lady &  \\
6  & from & on &  & when & its &  \\
7  & to & over &  & playwright & a &  \\
8  & , & , &  & a & Jack &  \\
9  & on & onto &  & his & them &  \\
10  & for & . &  & their & Henry &  \\
\hline
1  & \multicolumn{3}{c|}{\emph{playwright}} & \multicolumn{3}{c}{\emph{when}} \\
\hline
2  & duo & cast &  & When & . & \cellcolor[rgb]{0.843, 0.188, 0.153}{Julia} \\
3  & troupe & convict &  & as & , & \cellcolor[rgb]{0.843, 0.188, 0.153}{Jacques} \\
4  & theologian & audience &  & time & and & \cellcolor[rgb]{0.843, 0.188, 0.153}{Arthur} \\
5  & libretto & stage &  & about & ; & \cellcolor[rgb]{0.843, 0.188, 0.153}{George} \\
6  & inventor & couple &  & that & at &  \\
7  & Shakespeare & finale &  & at & in &  \\
8  & intruder & theatre &  & who & as &  \\
9  & author & principal &  & during & on &  \\
10  & theatre & production &  & other & but &  \\
\hline
1  & \multicolumn{3}{c|}{\emph{he}} & \multicolumn{3}{c}{\emph{took}} \\
\hline
2  & He & it & \cellcolor[rgb]{0.843, 0.188, 0.153}{she} & taken & arrived &  \\
3  & him & the & \cellcolor[rgb]{0.843, 0.188, 0.153}{Tech} & taking & heard &  \\
4  & his & his & \cellcolor[rgb]{0.843, 0.188, 0.153}{McCall} & take & came &  \\
5  & they & Lady &  & made & was &  \\
6  & who & she &  & finished & left &  \\
7  & the & they &  & won & met &  \\
8  & His & " &  & started & appeared &  \\
9  & was & asked &  & takes & reached &  \\
10  & she & a &  & went & had &  \\
\hline
1  & \multicolumn{3}{c|}{\emph{his}} & \multicolumn{3}{c}{\emph{bow}} \\
\hline
2  & him & the & \cellcolor[rgb]{0.843, 0.188, 0.153}{custody} & Bow & place & \cellcolor[rgb]{0.843, 0.188, 0.153}{anger} \\
3  & the & to &  & vow & last & \cellcolor[rgb]{0.843, 0.188, 0.153}{adopted} \\
4  & His & part &  & record & final & \cellcolor[rgb]{0.843, 0.188, 0.153}{chair} \\
5  & their & a &  & Drop & wife & \cellcolor[rgb]{0.988, 0.553, 0.349}{retirement} \\
6  & a & an &  & drop & leave & \cellcolor[rgb]{0.988, 0.553, 0.349}{hat} \\
7  & to & over &  & booth & post & \cellcolor[rgb]{0.784, 0.745, 0.565}{wife} \\
8  & in & off &  & memorial & passport &  \\
9  & its & place &  & chance & part &  \\
10  & her & it &  & snaps & first &  \\
\hline
\end{tabular}
\end{center}
\caption{Example top-k from bert-wwm-24, gpt2-774, and N-grams using the Wiki02 training partition at block 102 with ground-truth text `at the playwright when he took his bow'.  }
\end{table}

\begin{table}[H]  \label{tab::Wiki02-sample-582}
\begin{center}
\begin{tabular}{r|rrr|rrr}
Rank & BERT & GPT2 & Ngrams & BERT & GPT2 & Ngrams \\ \hline
1  & \multicolumn{3}{c|}{\emph{the}} & \multicolumn{3}{c}{\emph{offensive}} \\
\hline
2  & The & these & \cellcolor[rgb]{0.843, 0.188, 0.153}{an} & offense & opposing & \cellcolor[rgb]{0.843, 0.188, 0.153}{Ancients} \\
3  & THE & this &  & Offensive & shooter & \cellcolor[rgb]{0.843, 0.188, 0.153}{30} \\
4  & their & his &  & attacking & shot & \cellcolor[rgb]{0.843, 0.188, 0.153}{top} \\
5  & an & those &  & advancing & restricted & \cellcolor[rgb]{0.843, 0.188, 0.153}{oldest} \\
6  & a & a &  & Defensive & defending &  \\
7  & they & them &  & attack & three &  \\
8  & by & their &  & advance & players &  \\
9  & to & any &  & rushing & defender &  \\
10  & his & three &  & opposing & lane &  \\
\hline
1  & \multicolumn{3}{c|}{\emph{players}} & \multicolumn{3}{c}{\emph{violate}} \\
\hline
2  & Players & shooter & \cellcolor[rgb]{0.843, 0.188, 0.153}{,} & violating & are & \cellcolor[rgb]{0.843, 0.188, 0.153}{of} \\
3  & members & defender &  & violation & in & \cellcolor[rgb]{0.988, 0.553, 0.349}{are} \\
4  & player & team &  & abide & on &  \\
5  & men & or &  & violated & enter &  \\
6  & games & teams &  & comply & is &  \\
7  & musicians & offensive &  & obey & pass &  \\
8  & plays & possession &  & violations & drives &  \\
9  & Members & and &  & enforce & attempt &  \\
10  & soldiers & opponents &  & adherence & attempts &  \\
\hline
1  & \multicolumn{3}{c|}{\emph{the}} & \multicolumn{3}{c}{\emph{rule}} \\
\hline
2  & The & this & \cellcolor[rgb]{0.843, 0.188, 0.153}{natural} & rules & violation & \cellcolor[rgb]{0.843, 0.188, 0.153}{law} \\
3  & THE & these & \cellcolor[rgb]{0.843, 0.188, 0.153}{universal} & law & restricted & \cellcolor[rgb]{0.988, 0.553, 0.349}{constitution} \\
4  & it & any & \cellcolor[rgb]{0.843, 0.188, 0.153}{his} & Rules & NBA & \cellcolor[rgb]{0.784, 0.745, 0.565}{restoration} \\
5  & a & such & \cellcolor[rgb]{0.843, 0.188, 0.153}{not} & Rule & lane & \cellcolor[rgb]{0.784, 0.745, 0.565}{Nile} \\
6  & an & that &  & plan & shooter & \cellcolor[rgb]{0.784, 0.745, 0.565}{Azores} \\
7  & by & violation &  & decision & prohibited & \cellcolor[rgb]{0.784, 0.745, 0.565}{end} \\
8  & was & their &  & partner & law & \cellcolor[rgb]{0.784, 0.745, 0.565}{universe} \\
9  & this & an &  & leader & order & \cellcolor[rgb]{0.784, 0.745, 0.565}{season} \\
10  & its & NBA &  & road & opposing & \cellcolor[rgb]{0.784, 0.745, 0.565}{island} \\
\hline
1  & \multicolumn{3}{c|}{\emph{,}} & \multicolumn{3}{c}{\emph{no}} \\
\hline
2  & by & and & \cellcolor[rgb]{0.843, 0.188, 0.153}{then} & No & the & \cellcolor[rgb]{0.843, 0.188, 0.153}{experience} \\
3  & to & that & \cellcolor[rgb]{0.843, 0.188, 0.153}{of} & not & they &  \\
4  & of & by &  & all & a &  \\
5  & who & in &  & only & it &  \\
6  & for & of &  & little & an &  \\
7  & ( & ( &  & still & he &  \\
8  & that & or &  & just & or &  \\
9  & are & on &  & lot & such &  \\
10  & and & against &  & rest & their &  \\
\hline
\end{tabular}
\end{center}
\caption{Example top-k from bert-wwm-24, gpt2-774, and N-grams using the Wiki02 training partition at block 582 with ground-truth text `the offensive players violate the rule , no'.  }
\end{table}

\begin{table}[H]  \label{tab::Wiki02-sample-564}
\begin{center}
\begin{tabular}{r|rrr|rrr}
Rank & BERT & GPT2 & Ngrams & BERT & GPT2 & Ngrams \\ \hline
1  & \multicolumn{3}{c|}{\emph{tracked}} & \multicolumn{3}{c}{\emph{through}} \\
\hline
2  & Tracking & Island & \cellcolor[rgb]{0.843, 0.188, 0.153}{in} & Through & the & \cellcolor[rgb]{0.843, 0.188, 0.153}{near} \\
3  & track & had &  & across & into &  \\
4  & accelerating & was &  & faced & to &  \\
5  & ran & and &  & throughout & in &  \\
6  & overtook & is &  & Across & a &  \\
7  & tracking & County &  & hit & over &  \\
8  & trailed & , &  & though & slowly &  \\
9  & neared & has &  & crossed & on &  \\
10  & Track & Channel &  & visited & from &  \\
\hline
1  & \multicolumn{3}{c|}{\emph{Atlantic}} & \multicolumn{3}{c}{\emph{Canada}} \\
\hline
2  & Northeastern & the & \cellcolor[rgb]{0.843, 0.188, 0.153}{the} & Scotia & waters &  \\
3  & Nova & a &  & North & hurricane &  \\
4  & East & several &  & Canadian & Ocean &  \\
5  & New & an &  & Newfoundland & City &  \\
6  & Scotia & coastal &  & Maine & , &  \\
7  & eastern & some &  & States & and &  \\
8  & Southeast & to &  & Carolina & Cape &  \\
9  & Maine & , &  & China & Island &  \\
10  & to & this &  & Quebec & Islands &  \\
\hline
1  & \multicolumn{3}{c|}{\emph{,}} & \multicolumn{3}{c}{\emph{no}} \\
\hline
2  & did & and & \cellcolor[rgb]{0.843, 0.188, 0.153}{.} & No & the & \cellcolor[rgb]{0.843, 0.188, 0.153}{another} \\
3  & to & on & \cellcolor[rgb]{0.843, 0.188, 0.153}{and} & little & she & \cellcolor[rgb]{0.843, 0.188, 0.153}{which} \\
4  & as & in &  & there & it & \cellcolor[rgb]{0.843, 0.188, 0.153}{damage} \\
5  & had & to &  & any & Fran & \cellcolor[rgb]{0.843, 0.188, 0.153}{a} \\
6  & with & with &  & nothing & and & \cellcolor[rgb]{0.843, 0.188, 0.153}{attaining} \\
7  & also & the &  & not & however & \cellcolor[rgb]{0.843, 0.188, 0.153}{peaking} \\
8  & that & at &  & only & a & \cellcolor[rgb]{0.988, 0.553, 0.349}{the} \\
9  & was & for &  & none & this & \cellcolor[rgb]{0.988, 0.553, 0.349}{revealing} \\
10  & would & over &  & minimal & Ellen & \cellcolor[rgb]{0.988, 0.553, 0.349}{its} \\
\hline
1  & \multicolumn{3}{c|}{\emph{impact}} & \multicolumn{3}{c}{\emph{was}} \\
\hline
2  & affected & damage & \cellcolor[rgb]{0.843, 0.188, 0.153}{dialog} & is & on & \cellcolor[rgb]{0.843, 0.188, 0.153}{whatsoever} \\
3  & affect & significant & \cellcolor[rgb]{0.988, 0.553, 0.349}{structure} & wasn & from & \cellcolor[rgb]{0.843, 0.188, 0.153}{.} \\
4  & effect & major & \cellcolor[rgb]{0.988, 0.553, 0.349}{damage} & been & to & \cellcolor[rgb]{0.843, 0.188, 0.153}{on} \\
5  & impacts & flooding &  & were & or &  \\
6  & initial & other &  & to & of &  \\
7  & significant & storm &  & did & in &  \\
8  & impacted & direct &  & a & there &  \\
9  & initially & storms &  & would & upon &  \\
10  & aspect & damages &  & by & were &  \\
\hline
\end{tabular}
\end{center}
\caption{Example top-k from bert-wwm-24, gpt2-774, and N-grams using the Wiki02 training partition at block 564 with ground-truth text `tracked through Atlantic Canada , no impact was'.  }
\end{table}

\putbib
\end{bibunit}
\end{document}